\documentclass[11pt]{article}

\usepackage{genailab}
\usepackage{inconsolata}
\usepackage{graphicx}
\usepackage{booktabs}
\usepackage{makecell}
\usepackage{multirow}
\usepackage{xcolor}
\usepackage{amsmath}
\usepackage{amsfonts}
\usepackage{nicefrac}
\usepackage{enumitem}
\usepackage{float}
\usepackage{placeins}
\usepackage{stfloats}
\usepackage{fvextra}
\newcommand{\sigstars}[1]{\rlap{\textsuperscript{#1}}}
\newcommand{\metricpad}{\vphantom{\textsuperscript{***}}}
\newcommand{\skill}[1]{\texttt{\bfseries #1}}
\newcommand{\pvar}[1]{\texttt{\{#1\}}}
\definecolor{promptbg}{HTML}{FFEEE0}
\definecolor{promptframe}{HTML}{C8BE6A}
\newtcolorbox{promptfigurebox}{
  enhanced,
  colback=promptbg,
  colframe=promptframe,
  boxrule=1.1pt,
  arc=2.5mm,
  left=7pt,
  right=7pt,
  top=6pt,
  bottom=6pt,
  width=\textwidth
}
\begin{document}


\newcommand{\systemname}{\textsc{StatefulDiscovery}}


\paperheader{\today}

\papertitle{%
  {\color{sustechOrange}\systemname}:\,
  Evidence-Calibrated Claim Formation in Open-Ended Scientific Discovery
}

\paperauthors{%
  Jiayao Chen$^{1}$ \quad
  Shi Liu$^{1}$ \quad
  Linyi Yang$^{1,\dagger}$
}

\paperaffil{%
  $^{1}$\,Southern University of Science and Technology
}

\begin{paperabstract}
\noindent
Open-ended scientific discovery asks agents to move beyond executing analyses for predefined questions. Across multiple rounds of exploration, a discovery agent must decide which phenomena warrant investigation while avoiding overinterpretation, where emerging claims exceed the evidential scope of the analyses. This creates an evidence-calibration problem: the exploration trajectory must be coupled with claim status so that evidence can guide both what to investigate next and what can be claimed. We introduce \textsc{StatefulDiscovery}, a discovery framework that externalizes investigation state and uses it to coordinate frontier selection, evidence acquisition, and claim adjudication. We evaluate \textsc{StatefulDiscovery} across 40 real-data discovery tasks. Compared with several baselines, \textsc{StatefulDiscovery} produces more claims overall judged to be both well-supported and high-value. Ablations indicate distinct roles for structured hypotheses, local adjudication, frontier control and persistent states. Together, these results suggest that explicit discovery state can couple exploration with evidence-calibrated claim formation.

\codeline{\href{https://github.com/SUSTech-GenAI/StatefulDiscovery.git}{\texttt{https://github.com/SUSTech-GenAI/StatefulDiscovery.git}}}

\correspondenceline{\href{mailto:12531182@mail.sustech.edu.cn}{\texttt{12531182@mail.sustech.edu.cn}} \quad \href{mailto:yangly6@sustech.edu.cn}{\texttt{yangly6@sustech.edu.cn}}}
\end{paperabstract}

\daggernote

\section{Introduction}
\label{sec:intro}

Autonomous research agents increasingly automate complex scientific workflows, including literature synthesis, hypothesis generation, experiment planning, and paper writing~\cite{zhou2026autonomousagents,ghareeb2026multiagent,gottweis2026coscientist,lu2026endtoend,weng2025deepscientist,mitchener2025kosmos}. However, most existing systems operate within a goal-driven paradigm, anchored by predefined targets such as a specific disease mechanism or a benchmark metric. In contrast, open-ended data-driven discovery presents a more unconstrained challenge~\cite{agarwal2025autodiscovery}: given a dataset and broad scientific context, the agent must identify which phenomena warrant investigation. The challenge in this setting lies not merely in generating executable code for a specific analytical query, but in \emph{evidence calibration}, determining whether emerging claims are rigorously justified by the collected evidence. 

\begin{figure}[t]
    \centering
    \includegraphics[width=\linewidth,trim=30bp 0bp 23bp 0bp,clip]{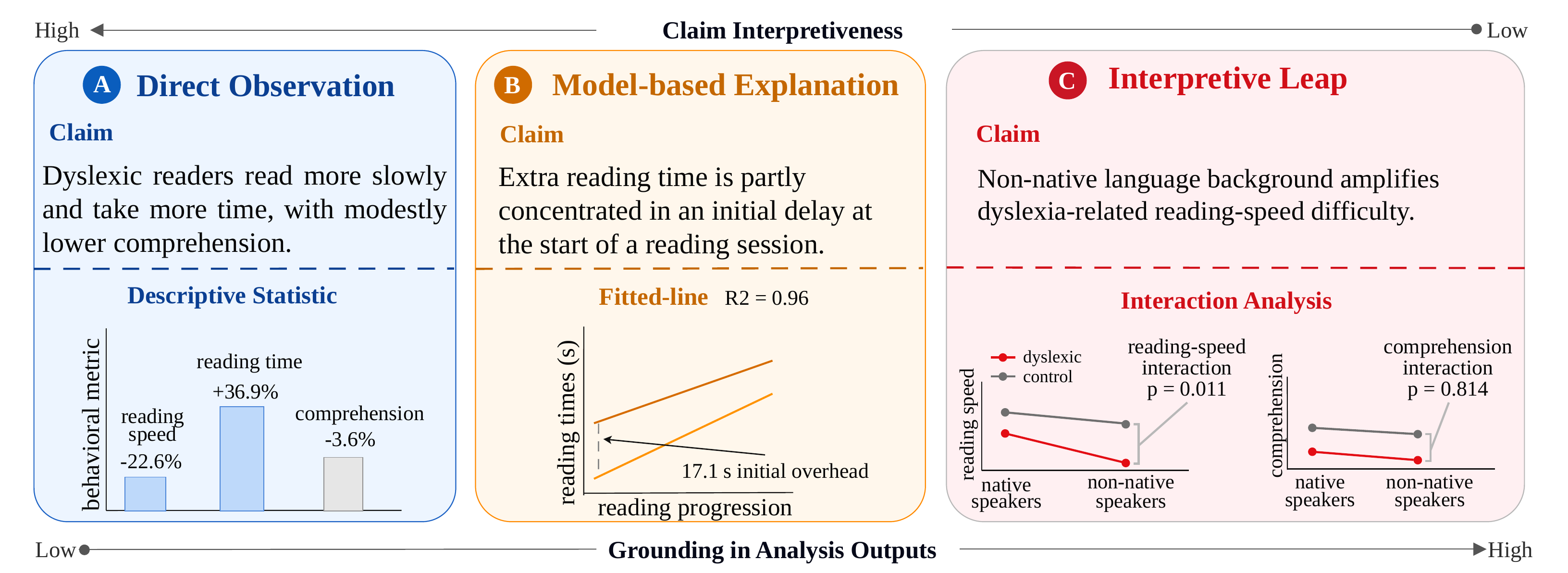}
    \caption{Support--interpretiveness trade-off in claim formation from data. (A) Descriptive claims stay close to observed metrics; (B) model-based explanations add interpretation while remaining tied to fitted evidence; (C) interpretive leaps arise when stronger claims exceed the evidential scope of the analyses.}
    \label{figure_1}
\end{figure}

This challenge exposes a fundamental tension in evidence-based claim formation: while highly interpretive claims can offer greater scientific value, they demand correspondingly stronger evidential grounding (Figure~\ref{figure_1}). For example, a descriptive claim about slower reading in dyslexic readers is directly supported by observed metrics, while a model-fitted explanation of initial reading overhead adds interpretation and remains tied to an empirical phenomenon in the dataset~\cite{gu2024blade}. However, an agent risks an \emph{interpretive leap} if it asserts that a non-native language background amplifies this difficulty without establishing comparable task constraints across groups. We use \emph{overinterpretation}~\cite{boutron2010reporting,mcgrath2017overinterpretation} to define the failure mode where the semantic strength of an agent's claim exceeds the empirical scope of its supporting analyses.

In autonomous open-ended discovery, overinterpretation is a critical vulnerability. Prior systems provide mechanisms for search, optimization, or retrospective validation~\citep{agarwal2025autodiscovery,novikov2025alphaevolve,yan2026pacevolve}, yet they leave a crucial evidence-calibration problem structurally unresolved: \textbf{how the exploration trajectory should be coupled with the status of emerging claims?} Consequently, agents relying on implicit working memory or unstructured context windows often lose track of weak evidence, discard unresolved alternatives, and overlook evidence gaps. The necessary insight is to introduce an explicit \emph{epistemic state}, allowing the agent to keep weakly supported claims provisional, preserve competing hypotheses, and explicitly track evidence gaps across iterations.

To this end, we present \systemname{}, an autonomous discovery framework governed by an explicit epistemic state. The agent externalizes the evolving investigation state as persistent objects tracking observations, active investigations, collected evidence, and the evidential status of candidate claims. It regulates exploration through a dual-layer architecture: \textit{global frontier control} selects the next direction to investigate, while \textit{local hypothesis-set adjudication} compares competing hypotheses and calibrates candidate claims against the executed evidence. Together, these mechanisms systematically link frontier selection, evidence acquisition, and evidence-calibrated claim formation into a unified process.

We evaluate \systemname{} on 40 real-data tasks covering biomedical, social-science, behavioral, and cross-domain datasets. To assess performance under a finite exploration budget, we introduce a novel claim-level evaluation protocol that explicitly decouples evidential support from scientific discovery value. Our empirical results demonstrate that \systemname{} yields claim sets that are preferred in pairwise comparisons. Overall, it produces 23\% more high-quality claims than the closest baseline, while also being preferred over that baseline in 31 of 40 pairwise comparisons. Ablation studies further confirm that structured hypothesis sets improve discovery value, while the dual-layer adjudication and control mechanisms optimize the trade-off between exploration yield and claim validity.

We evaluate \systemname{} on 40 real-data tasks covering biomedical, social-science, behavioral, and cross-domain datasets. To assess performance under a finite exploration budget, we introduce a novel claim-level evaluation protocol that explicitly decouples evidential support from scientific discovery value. Our empirical results demonstrate that \systemname{} yields claim sets that are preferred in pairwise comparisons. Overall, it produces 6.25pp more high-quality claims than the closest baseline, while also being preferred over that baseline in 31 of 40 pairwise comparisons. Ablation studies further confirm that structured hypothesis sets improve discovery value, while the dual-layer adjudication and control mechanisms optimize the trade-off between exploration yield and claim validity. Figure~\ref{fig:bix52_complete_case} (Appendix ~\ref{app:complete_trace_example}) gives an example of  one complete trace from the a bioinformatic task.

To our knowledge, we are the first to utilize the discovery state as a control interface for open-ended exploration. Under this framework, the status of emerging claims actively directs the agent's subsequent investigations and calibrates the confidence of its conclusions.
\systemname{} implements this through persistent state updates that mediate coupled decisions about frontier selection, evidence acquisition, and claim assessment across discovery rounds. 
We instantiate and evaluate this design in open-ended discovery tasks and introduce a claim-level evaluation protocol that separates evidential support from discovery value and compares claim formation under a finite budget.

\section{Related Work}
\subsection{Evolutionary Research Agents}
Autonomous research systems accelerate the scientific process~\cite{zhou2026autonomousagents,lu2026endtoend,ghareeb2026multiagent,gottweis2026coscientist}. Frameworks such as Kosmos~\cite{mitchener2025kosmos}, DeepScientist~\cite{weng2025deepscientist}, and AutoSOTA~\cite{li2026autosota} further coordinate long-horizon discovery around shared contexts or benchmark targets. In parallel, evolutionary search agents iteratively propose and select candidate program variants, objectives, or computable scorers based on explicit evaluator feedback~\cite{yan2026pacevolve,novikov2025alphaevolve,du2025saga}. However, both paradigms constrain discovery by requiring either a supplied benchmark query, a fixed target, or a scalar fitness function. To bypass these dependencies, we focus on open-ended claim formation when no specific analytical query, target claim, or optimization goal is provided in advance.

\subsection{Agent Memory Management}
Memory management methods help long-horizon agents store, retrieve, and organize task context through workflows, working memory, traces, or learned memory policies~\cite{wu2024stateflow,fourney2024magenticone,hu2024hiagent,wang2024awm,yu2026agenticmemory}. BeliefMem~\cite{liao2026beliefmem} further argues that memory should retain multiple candidate conclusions under partial observability. These methods primarily preserve facts, experiences, or execution trajectories to support predefined tasks or later experience reuse. StatefulDiscovery uses persistence for a more specific scientific role. The external state serves as a control interface for evidence-calibrated claim formation, not as a general memory mechanism.

\subsection{Scientific Discovery Control}
Existing systems guide scientific exploration in more specified settings. AutoDiscovery~\cite{agarwal2025autodiscovery} prioritizes hypotheses with Bayesian surprise and tree search. PiFlow~\cite{pu2025piflow} and PiEvo~\cite{pu2026pievo} use scientific principles and uncertainty reduction. EXPERIGEN~\cite{sengupta2026experigen} couples hypothesis generation with experimental validation, while HypoExplore~\cite{koo2026hypoexplore} maintains hypothesis memory and confidence updates for visual architecture discovery. These systems typically assume a hypothesis space, validation target, scientific principle, human-specified direction, or external feedback signal. \textsc{StatefulDiscovery} instead asks how accumulated evidence should guide both subsequent investigation and claim strength without a target claim or a scoring criterion in advance.

Other work develops infrastructure for autonomous science, including federated workflow agents~\cite{pauloski2025academy}, cooperative closed-loop discovery systems~\cite{pauloski2025agenticdiscovery}, governed drug-discovery agents~\cite{cao2026mozi}, and auditable research frameworks~\cite{rasheed2026auditability}. These systems can coordinate, govern, or audit discovery workflows, but they do not by themselves specify how evidence status should drive exploration and claim formation. \textsc{StatefulDiscovery} targets this discovery-control layer by making discovery state an interface between evidential status, frontier selection, and claim formation.

\section{\systemname{}: Evidence-Calibrated Claim Formation}
We define \textsc{StatefulDiscovery} as a framework that maintains an explicit, evolving investigation state during open-ended scientific discovery. It organizes the process around bounded investigations, competing hypotheses, executable evidence, and evidence-calibrated claims. This state makes local claim formation visible to global exploration control. Figure~\ref{fig:method_overview} provides an overview of our agent as instantiated for data discovery.

\subsection{Epistemic State and Discovery Objects}
\begin{figure}[t]
    \centering
    \includegraphics[width=\textwidth]{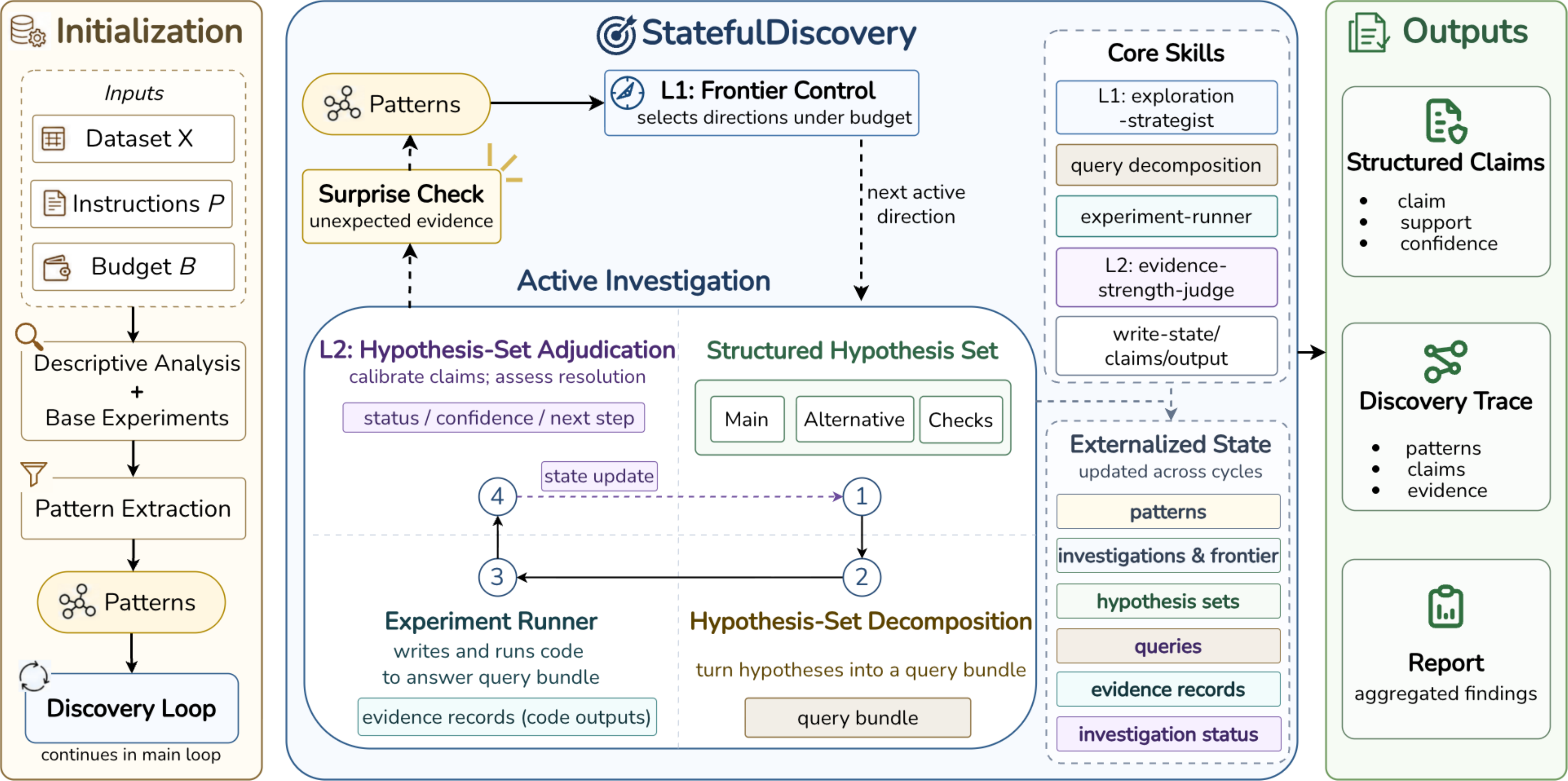}
    \caption{Overview of \systemname{}. (a) Initialization uses the dataset, instructions, and budget to surface and prioritize candidate patterns. (b) The framework externalizes discovery state as persistent objects in a stateful loop: under L1 frontier control, the agent selects a direction; within an active investigation, it maintains hypotheses, runs executable analyses, and uses L2 adjudication to evaluate evidence, calibrate claim confidence, assess investigation status, and identify next steps. (c) The agent returns structured claims, a discovery trace, and an aggregated report.}
    \label{fig:method_overview}
\end{figure}

The framework represents discovery state through seven persistent discovery objects that the agent reads and updates throughout discovery. These objects correspond to the externalized-state block in Figure~\ref{fig:method_overview} and provide the interface between initialization, the active discovery loop, and final reporting. This design reflects a process view of scientific discovery: phenomena motivate hypotheses, hypotheses guide tests, and evidence updates what can be claimed~\cite{whewell1840inductive,klahr1988dual,langley1987scientific,darden2002strategies}.

A \textbf{pattern} is an observed phenomenon that may warrant further investigation. An \textbf{investigation} is a bounded line of exploration over one or more related patterns. Within each investigation, a \textbf{structured hypothesis set} includes a main hypothesis, alternatives, artifact checks, and robustness checks. A \textbf{query} is an executable analysis designed to discriminate among hypotheses, and its outputs become \textbf{evidence} records that update hypothesis and claim status. An \textbf{investigation status} summarizes what has been resolved under available evidence, while a \textbf{frontier state} records what remains worth exploring under the remaining budget.

\begin{center}
\begingroup
\setlength{\fboxrule}{0.35pt}
\setlength{\fboxsep}{3pt}
\fbox{\begin{minipage}{0.96\linewidth}
\footnotesize
\textbf{Example discovery state.}\\
\textbf{Pattern:} faster reading is associated with a larger accuracy drop in dyslexic than in controls;\\
\textbf{Investigation:} test vulnerability in dyslexic readers to speed-induced accuracy loss; \\
\textbf{Structured hypothesis set:} main speed--accuracy interaction, with page-confounding, sample-size, and speed-distribution checks;\\
\textbf{Query:} fit a dyslexia $\times$ speed model, test page interactions, bootstrap the interaction, and compare speed distributions; \\
\textbf{Evidence:} significant interaction and no page confounding, but inconclusive bootstrap robustness; \\
\textbf{Investigation status:} partially supported but robustness-limited, so the claim remains cautious; \\
\textbf{Frontier state:} carry forward a cautious claim and leave the broader reading mechanism open for later investigation.
\end{minipage}}
\endgroup
\end{center}

\subsection{Discovery Loop and Dual-layer Architecture}
The discovery loop uses the externalized state in a dual-layer architecture. As shown in Figure~\ref{fig:method_overview}, the agent uses L1 frontier control to decide where to allocate the next unit of budget. Within the active investigation, it uses L2 local hypothesis-set adjudication to evaluate evidence, calibrate hypothesis and claim status, and update the externalized state. Before any investigation is active, initialization runs exploratory data analysis and low-cost base analyses to surface candidate patterns. The agent assigns each observed pattern an initial priority score that reflects its relative priority for further investigation. These prioritized patterns seed the initial frontier.

At the frontier level, the agent consults the externalized state to allocate effort across the broader discovery process. This L1 decision reads the existing pattern priorities together with active investigation status, frontier saturation, red flags, and remaining budget. Based on this state, the agent opens an investigation, deepens the current one, shifts to another direction, retires an unpromising investigation, or stops. Resolved investigations release budget toward high-priority patterns, while unresolved evidence gaps keep the agent focused on the current one. When the frontier is saturated or remaining directions appear driven by artifacts, the agent may retire them or stop.

Once the agent activates an investigation through L1 frontier control, the numbered steps in Figure~\ref{fig:method_overview} describe the recurrent structure of the active investigation. The agent maintains a structured hypothesis set over the active direction, including a main hypothesis, alternatives, artifact checks, and robustness checks. It then translates this hypothesis set into an executable query bundle, runs the corresponding analyses, and records code outputs as evidence.

At the local level, the agent uses L2 adjudication to evaluate evidence for the active investigation. It checks whether the executed evidence answers the intended query, compares the main hypothesis against alternatives, artifact checks, and robustness checks, and assigns each hypothesis an outcome: support, weaken, refute, or inconclusive. It also records an investigation-level assessment with the best-supported explanation, remaining concerns, resolution confidence, red flags, and recommendations for next steps. This update guides whether the investigation should yield a claim, be deepened or revised, or return to frontier control.

During L2 adjudication, the agent also records a surprise signal for each executed query by comparing the query's \texttt{expected\_effect}, written during hypothesis-set decomposition, with the observed evidence returned by the experiment runner. This signal is heuristic and is not a calibrated statistical test. It indicates whether the observed result contradicts or substantially departs from the expected effect. If the investigation is otherwise resolved, a surprising result can be promoted to a new pattern for the next L1 frontier decision; if the investigation remains unresolved, the signal is retained as a concern or cue in the current L2 assessment.

\subsection{Skill-Based Implementation}
\label{sec:skill_interfaces}
In our implementation, skills provide the operational interface between the discovery agent and the externalized state. After initialization, each cycle follows a fixed sequence: \textit{L1 frontier selection $\rightarrow$ investigation decomposition $\rightarrow$ experiment execution $\rightarrow$ L2 adjudication $\rightarrow$ state update $\rightarrow$ next L1 decision}. The discovery loop is specified by fixed skill interfaces, without a learned policy or a hand-coded scoring function. Each skill specifies which state records to read, what reasoning or analysis procedure to perform, and what structured updates to write back, following the agent-skills abstraction~\cite{jiang2026agenticskills,shen2026skillfoundry}.

The L1 frontier-control skill reads pattern priorities, active-investigation status, red flags, frontier saturation, and remaining budget, and returns one frontier action from a fixed set: create, attach, deepen, switch, retire, or stop. The L2 adjudication skill reads the active hypothesis set, query bundle, and executed evidence, and writes hypothesis outcomes, investigation confidence, red flags, resolution status, recommendations for next steps, surprise signals, and finalizable claims. The full skill interfaces, state schemas, and expanded read--write protocol are provided in Appendix~\ref{app:skill_all}. Figure~\ref{fig:l1_l2_schema} gives the L1/L2 operational view.

\section{Experiments}

\subsection{Experimental Setup}
We evaluate \textsc{StatefulDiscovery} on 40 open-ended data-discovery tasks constructed from existing question-answering benchmarks. For each source task, we retain the dataset together with the minimal file and variable information needed to understand and parse the files. We remove the benchmark-specific question, target answer, and concrete analytical query. With only file paths, available variable descriptions, and a fixed budget as input, each agent is instructed to produce evidence-supported claims without a target claim to verify.

The evaluation set contains 21 tasks from BixBench\footnote{\url{https://huggingface.co/datasets/phylobio/BixBench-Verified-50}}~\cite{mitchener2025bixbench}, 13 from BLADE~\cite{gu2024blade}, and 6 from DiscoveryBench~\cite{majumder2025discoverybench}, covering biomedical, social-science, behavioral, and cross-domain datasets. BLADE and DiscoveryBench follow the benchmark sources used by AutoDiscovery\cite{agarwal2025autodiscovery}, while BixBench adds a computational-biology setting. Unless otherwise stated, LLM calls use Qwen3.5-plus. All agents run without web search or external literature retrieval and receive a budget of 40 code executions per task. Additional dataset details, sandbox setup, and instruction templates are provided in Appendix~\ref{app:experimental_details}.

\subsection{Evaluation Metrics}
Open-ended discovery requires evaluating both whether a claim is evidentially supported and whether it adds useful understanding~\cite{whewell1840inductive,darden2002strategies}. We therefore decouple claim evaluation into two axes. \emph{Evidential Support} (ES) measures a evidence-to-claim alignment, whether the executed analysis output supports a claim. \emph{Discovery Value} (DV) measures whether the claim adds nontrivial explanatory or structural understanding beyond a fixed first-pass analysis for the same task. Both metrics are scored on a 1--5 scale using rubric-based LLM judging~\cite{liu2023geval,zheng2023judging,dubois2024length}. Formally, let \(\rho_{\mathrm{ES}}\) and \(\rho_{\mathrm{DV}}\) denote the ES and DV judge prompts, respectively. For task \(i\), claim \(c_{ij}\), associated evidence \(e_{ij}\), dataset description \(d_i\), and first-pass analysis results \(f_i\), we compute
\begin{align}
    s^{\mathrm{ES}}_{ij} &= \mathrm{LLM}(\rho_{\mathrm{ES}}; d_i, c_{ij}, e_{ij}), \\
    s^{\mathrm{DV}}_{ij} &= \mathrm{LLM}(\rho_{\mathrm{DV}}; d_i, c_{ij}, f_i),
\end{align}
where \(s^{\mathrm{ES}}_{ij},s^{\mathrm{DV}}_{ij} \in \{1,2,\ldots,5\}\).

Before judging, outputs from all methods are normalized into a common claim--evidence format. For each claim, we extract the claim text and its supporting evidence or analysis output, excluding method-specific intermediate artifacts. To avoid favoring systems that produce more claims, ES and DV are averaged first within each task and then across tasks. The two dimensions distinguish different failure modes. A claim that is accurate but only restates descriptive statistics may have high ES and low DV, whereas a strongly explanatory but insufficiently supported claim may have high DV and low ES. Therefore, our  main composite HQ metric counts a claim only when it satisfies both ES $\ge 4$ and DV $\ge 4$. We also compare each method's normalized final claim set pairwise, using position-swapped judging and treating disagreements as ties. All LLM-based judges use Gemini-3.1-pro, with prompt templates provided in Appendix~\ref{app:eval_prompts}.

\subsection{Baselines}
We compare \textsc{StatefulDiscovery} against four baselines that can operate under the same open-ended data-discovery protocol: a raw discovery agent, AutoDiscovery~\cite{agarwal2025autodiscovery}, OpenEvolve\footnote{\href{https://github.com/algorithmicsuperintelligence/openevolve}{OpenEvolve repository}}, and SAGA~\cite{du2025saga}. To reduce prompt-induced confounds, all methods receive the same dataset files, available variable descriptions, budget, and open-ended instruction to produce evidence-supported claims. Because directly comparable open-ended baselines are limited, we adapt each method to this input--output protocol while preserving its core mechanism. The baseline instructions and adaptation prompt templates are provided in Appendix~\ref{app:baseline_prompts}. All agents use the same Qwen3.5-plus backbone.

\begin{itemize}[leftmargin=*,itemsep=1pt,topsep=1pt,parsep=0.5pt,partopsep=0.5pt]
    \item \textbf{Raw agent} is a direct coding-agent baseline that uses the same backbone model and discovery environment as \textsc{StatefulDiscovery}, but receives only an analysis-and-report instruction. It does not use persistent discovery objects, L1 frontier control, or L2 adjudication.

    \item \textbf{AutoDiscovery}~\citep{agarwal2025autodiscovery} is a prior open-ended discovery method based on Bayesian surprise from LLM samples and MCTS. We use the authors' open-source implementation.

    \item \textbf{OpenEvolve} is an open-source evolutionary coding framework based on the AlphaEvolve paradigm~\citep{novikov2025alphaevolve}. We adapt it to open-ended discovery by cold-starting each run with an initial discovery direction and executable analysis program. Before each code mutation, the model proposes a concrete hypothesis or analysis direction, and the mutated program is scored by an LLM judge for output validity, claim alignment, and non-triviality.
    
    \item \textbf{SAGA}~\citep{du2025saga} evolves operational objectives and scorers for scientific optimization. In our setting, candidates are evidence-backed findings produced by executable analyses instead of directly mutable objects such as molecules or DNA sequences. We therefore implement candidate evolution at the analysis level: at each iteration, SAGA proposes discovery objectives, constructs or selects scorers, and scores the current finding population. High-scoring findings condition subsequent analyses, while newly generated findings are merged into the population and re-ranked for the next iteration.
\end{itemize}

\section{Results}
Table~\ref{tab:current_results} summarizes the main comparison on the 40-task evaluation set, and Figure~\ref{fig:claim_score_distributions} shows the corresponding claim-level score distributions. \textsc{StatefulDiscovery} achieves the highest overall Discovery Value ($3.09$) and the highest observed high-quality claim count and rate ($64$ claims, $24.5\%$), while SAGA is the closest baseline on HQ ($52$ claims, $20.8\%$).

\begin{table*}[t]
    \centering
    \small
    \caption{Main results on the 40-task open-ended discovery evaluation set. ES/DV are task-level means with standard deviations. Claims denotes the number of final claims. HQ reports claims with both $ES\ge4$ and $DV\ge4$ as count/rate. Bold marks the best value in each block; ES/DV stars denote one-sided paired Wilcoxon significance against all other agents in the block, using the weakest pairwise comparison (${}^{*}p<0.05$, ${}^{**}p<0.01$, ${}^{***}p<0.001$).}
    \label{tab:current_results}
    \begin{tabular*}{\linewidth}{@{\extracolsep{\fill}}llcccr@{}}
        \toprule
        Task source & Agent & ES $\uparrow$ & DV $\uparrow$ & Claims & HQ $\uparrow$ \\
        \midrule
        \multirow{5}{*}{\makecell[l]{BixBench\\\citep{mitchener2025bixbench}}} & Raw agent & $4.27 \pm 0.55$\metricpad & $2.51 \pm 0.59$\metricpad & 212 & $29/13.7\%$ \\
        & AutoDiscovery & $3.35 \pm 1.17$\metricpad & $2.47 \pm 0.66$\metricpad & 198 & $18/9.1\%$ \\
        & OpenEvolve & \textbf{4.74} $\pm$ \textbf{0.42}\sigstars{**}\metricpad & $1.31 \pm 0.39$\metricpad & 192 & $1/0.5\%$ \\
        & SAGA & $3.96 \pm 0.75$\metricpad & $2.86 \pm 0.95$\metricpad & 109 & $\textbf{37}/\textbf{33.9\%}$ \\
        & Ours & $3.75 \pm 0.76$\metricpad & \textbf{3.14} $\pm$ \textbf{0.89}\metricpad & 136 & $33/24.3\%$ \\
        \midrule
        \multirow{5}{*}{\makecell[l]{BLADE\\\citep{gu2024blade}}} & Raw agent & $4.11 \pm 0.64$\metricpad & $2.33 \pm 0.35$\metricpad & 115 & $11/9.6\%$ \\
        & AutoDiscovery & $4.42 \pm 1.04$\metricpad & $2.63 \pm 0.47$\metricpad & 130 & $11/8.5\%$ \\
        & OpenEvolve & \textbf{4.80} $\pm$ \textbf{0.34}\metricpad & $1.44 \pm 0.41$\metricpad & 91 & $3/3.3\%$ \\
        & SAGA & $3.48 \pm 1.11$\metricpad & $2.37 \pm 0.85$\metricpad & 96 & $11/11.5\%$ \\
        & Ours & $3.95 \pm 0.92$\metricpad & \textbf{2.86} $\pm$ \textbf{0.60}\metricpad & 86 & $\textbf{22}/\textbf{25.6\%}$ \\
        \midrule
        \multirow{5}{*}{\makecell[l]{DiscoveryBench\\\citep{majumder2025discoverybench}}} & Raw agent & $4.26 \pm 0.31$\metricpad & $2.61 \pm 0.67$\metricpad & 54 & $7/13.0\%$ \\
        & AutoDiscovery & $3.39 \pm 1.40$\metricpad & $2.90 \pm 0.29$\metricpad & 59 & $7/11.9\%$ \\
        & OpenEvolve & \textbf{4.91} $\pm$ \textbf{0.12}\metricpad & $1.92 \pm 0.55$\metricpad & 68 & $\textbf{12}/17.6\%$ \\
        & SAGA & $3.90 \pm 1.31$\metricpad & $2.59 \pm 0.37$\metricpad & 45 & $4/8.9\%$ \\
        & Ours & $3.41 \pm 1.14$\metricpad & \textbf{3.39} $\pm$ \textbf{0.41}\sigstars{*}\metricpad & 39 & $9/\textbf{23.1\%}$ \\
        \midrule
        \multirow{5}{*}{Overall} & Raw agent & $4.22 \pm 0.54$\metricpad & $2.47 \pm 0.53$\metricpad & 381 & \mbox{$47/12.3\%$} \\
        & AutoDiscovery & $3.70 \pm 1.24$\metricpad & $2.59 \pm 0.57$\metricpad & 387 & \mbox{$36/9.3\%$} \\
        & OpenEvolve & \textbf{4.78} $\pm$ \textbf{0.36}\sigstars{***}\metricpad & $1.44 \pm 0.46$\metricpad & 351 & \mbox{$16/4.6\%$} \\
        & SAGA & $3.80 \pm 1.04$\metricpad & $2.66 \pm 0.91$\metricpad & 250 & \mbox{$52/20.8\%$} \\
        & Ours & $3.76 \pm 0.87$\metricpad & \textbf{3.09} $\pm$ \textbf{0.76}\sigstars{*}\metricpad & 261 & \mbox{$\textbf{64}/\textbf{24.5\%}$} \\
        \bottomrule
    \end{tabular*}
\end{table*}

\subsection{Open-ended Scientific Discovery}
\begin{figure}[t]
    \centering
    \includegraphics[width=0.4\linewidth]{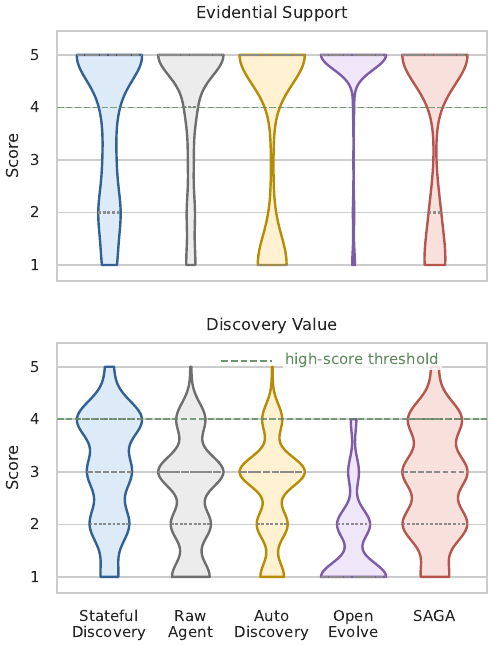}
    \caption{Claim-level score distributions across all 40 tasks. The dashed line marks the high-score threshold used in the HQ metric (${\mathrm{ES}}\ge4$ and ${\mathrm{DV}}\ge4$).}
    \label{fig:claim_score_distributions}
\end{figure}

\textsc{StatefulDiscovery} also produces fewer claims than the raw agent and AutoDiscovery, suggesting that its support--value gains are not simply due to larger output volume. OpenEvolve illustrates why separating ES from DV is useful: it obtains the highest ES ($4.78$), but the lowest DV ($1.44$) and HQ rate ($4.6\%$), with Figure~\ref{fig:claim_score_distributions} showing that its discovery-value scores concentrate below the high-score threshold.

A coarse manual inspection of OpenEvolve's final claims supports this interpretation. Of its 351 deduplicated claims, 74 (21.1\%) are data-availability or format-check statements. Removing these statements changes the overall scores only modestly, from ES $=4.78$ and DV $=1.44$ to ES $=4.77$ and DV $=1.61$, indicating that the low DV is not mainly driven by file- or format-level claims. Among the remaining claims, 189 of 277 (68.2\%) are simple pairwise associations or descriptive/statistical summaries, suggesting a broader tendency toward evidence-aligned but local statistical claims.

Table~\ref{tab:pairwise_results} provides a complementary claim-set evaluation. \textsc{StatefulDiscovery} is preferred over the raw agent on 25 of 40 tasks, over AutoDiscovery on 32, over OpenEvolve on all 40, and over SAGA on 31. This complements the claim-level metrics. SAGA is strong on individual claims, especially on BixBench, but pairwise evaluation favors \textsc{StatefulDiscovery} when judging the final claim set as a whole, including whether claims are mutually consistent and calibrated to the evidence. A GPT-5.5 second-judge check preserves the same qualitative pattern (Appendix~\ref{app:pairwise_robustness}), and a length analysis indicates that the gains are not explained by longer claim-set presentations (Appendix~\ref{app:pairwise_length}). Besides, Appendix~\ref{app:efficiency_cost} reports efficiency and cost, and Appendix~\ref{app:hurricane_case} provides a qualitative case study.

\begin{table}[t]
    \centering
    \small
    \caption{Pairwise judge preferences on the shared 40-task evaluation set.}
    \label{tab:pairwise_results}
    \begin{tabular*}{0.8\linewidth}{@{\extracolsep{\fill}}lccc@{}}
        \toprule
        Comparison & Win & Tie & Loss \\
        \midrule
        Ours vs. Raw agent & 25 & 6 & 9 \\
        Ours vs. AutoDiscovery & 32 & 6 & 2 \\
        Ours vs. OpenEvolve & 40 & 0 & 0 \\
        Ours vs. SAGA & 31 & 7 & 2 \\
        \bottomrule
    \end{tabular*}
\end{table}

\subsection{Human Validation of Automatic Scores}
We further validate the automatic claim-level scores against human annotations. For each metric, we sample 120 claims using stratified sampling over automatic judge scores and the five evaluated systems. The sample is balanced across systems and distributed across tasks to avoid concentrating the validation set on a small number of datasets. Two PhD-level annotators from different institutions, with backgrounds in statistics and biology, independently score the sampled claims using the same 1--5 rubrics. Table~\ref{tab:human_metric_validation} reports within-one agreement, mean absolute error, and rank correlation for human--human agreement and for the automatic judge against the mean human score. The automatic judge shows reasonable consistency with expert annotations, supporting its use as a scalable proxy for aggregate system-level comparisons.

\begin{table}[htbp]
    \centering
    \small
    \setlength{\tabcolsep}{2pt}
    \renewcommand{\arraystretch}{1.12}
    \caption{Human validation of automatic claim-level scores. Agreement is reported for Discovery Value (DV) and Evidential Support (ES) using within-one agreement, mean absolute error (MAE), and Spearman rank correlation.}
    \label{tab:human_metric_validation}
    \begin{tabular*}{0.8\linewidth}{@{\extracolsep{\fill}}llccc@{}}
        \toprule
        Metric & Comparison & \makecell{Within 1} & MAE & \makecell{Spear.\\$\rho$} \\
        \midrule
        DV & \makecell[l]{Human A vs. Human B} & $98.3\%$ & $0.233$ & $0.860$ \\
        DV & \makecell[l]{Human mean vs. Judge} & $91.7\%$ & $0.617$ & $0.782$ \\
        ES & \makecell[l]{Human A vs. Human B} & $96.6\%$ & $0.389$ & $0.809$ \\
        ES & \makecell[l]{Human mean vs. Judge} & $86.4\%$ & $0.627$ & $0.687$ \\
        \bottomrule
    \end{tabular*}
\end{table}

\subsection{Ablation Study}
Table~\ref{tab:ablation_results} isolates the framework components in a cumulative ablation. Pattern recognition adds persistent pattern records, priorities, and priority-ordered follow-up experiments. The structured-hypothesis-set variant groups related patterns into bounded investigations with main, alternative, artifact, and robustness roles, and decomposes each investigation into executable query bundles, but does not adjudicate claim status. L2 adjudication adds local evidence-strength judging and per-claim status updates for each active investigation. The full system adds L1 frontier control over the existing externalized state for selecting whether to create, deepen, switch, retire, or stop investigations. Because these components are sequentially dependent, each row adds one component to the previous configuration. Every row is an independent experiment run under the same tasks and execution budget.

\begin{table}[htbp]
    \centering
    \small
    \setlength{\tabcolsep}{2pt}
    \renewcommand{\arraystretch}{1.08}
    \caption{Cumulative ablation results. ES and DV are task-level means with standard deviations. Claims and HQ follow the same definitions as Table~\ref{tab:current_results}; the Raw agent and StatefulDiscovery rows reuse the Overall rows from the main comparison.}
    \label{tab:ablation_results}
    \begin{tabular*}{0.8\linewidth}{@{\extracolsep{\fill}}lcccc@{}}
        \toprule
        Method & ES $\uparrow$ & DV $\uparrow$ & Claims & HQ $\uparrow$ \\
        \midrule
        Raw agent & $\textbf{4.22} \pm \textbf{0.54}$ & $2.47 \pm 0.53$ & 381 & $47/12.3\%$ \\
        + Pattern & $3.52 \pm 1.04$ & $2.63 \pm 0.64$ & 283 & $39/13.8\%$ \\
        + Hypothesis-set & $3.18 \pm 1.56$ & $\textbf{3.23} \pm \textbf{0.71}$ & 277 & $62/22.4\%$ \\
        + L2 & $3.63 \pm 1.33$ & $3.11 \pm 1.02$ & 142 & $38/\textbf{26.8\%}$ \\
        \makecell[l]{+ L1 (ours)} & $3.76 \pm 0.87$ & $3.09 \pm 0.76$ & 261 & $\textbf{64}/24.5\%$ \\
        \bottomrule
    \end{tabular*}
\end{table}

The ablation highlights the trade-off among support, value, and yield. The raw agent obtains the highest ES, but its DV and HQ rate remain low, suggesting that evidence-aligned claims are often shallow. Pattern and hypothesis-set structure create more valuable investigation directions: the structured-hypothesis variant reaches the highest mean DV, but also lowers ES because more interpretive claims require calibration. L2 adjudication then filters unresolved or weakly supported claims, raising the HQ rate while reducing output volume. Adding L1 frontier control restores exploration yield under the same adjudication layer, maintaining similar ES and DV to the L2-only variant while producing the largest number of high-quality claims. 

\paragraph{Persistence States Ablation.}
The cumulative ablation above does not isolate externalized persistence. We therefore add a context-only condition that retains the same setting of full StatefulDiscovery, but remove the requirements and skill interfaces for maintaining cross-round patterns, investigations, and epistemic-state records. L1/L2 outputs remain available in the conversation context but are not written to external state for retrieval in later rounds. As shown in table \ref{tab:persistence_ablation}, removing the externalized state does not reduce DV, and it is slightly higher. The main degradation is instead in ES (\text{-}0.65) and HQ rate (\text{-}11.5 pp), while the HQ count falls from 64 to 24. This pattern is consistent with our proposed mechanism: context-only L1/L2 can still identify directions with explanatory value, but it is less able to carry forward the evolving evidential state and use it to constrain subsequent claim formation. Its HQ rate consequently falls close to that of the Raw Agent. These results suggest that the primary role of externalized state is to preserve discovery state and evidential constraints across rounds and bring them into subsequent exploration.

\begin{table}[t]
    \centering
    \footnotesize
    \setlength{\tabcolsep}{3pt}
    \caption{Matched ablation of externalized cross-round state. The context-only condition retains the functional L1/L2 roles but does not write or retrieve external discovery-state records. ES and DV are task-level means.}
    \label{tab:persistence_ablation}
    \begin{tabular*}{\linewidth}{@{\extracolsep{\fill}}lcccc@{}}
        \toprule
        Method & ES $\uparrow$ & DV $\uparrow$ & Claims & HQ $\uparrow$ \\
        \midrule
        Raw agent & $4.22$ & $2.47$ & 381 & $47/12.3\%$ \\
        Context-only & $3.11$ & $3.27$ & 184 & $24/13.0\%$ \\
        Full system & $3.76$ & $3.09$ & 261 & $\mathbf{64}/\mathbf{24.5\%}$ \\
        \bottomrule
    \end{tabular*}
\end{table}

\subsection{Backbone Model Sensitivity Analysis}
\begin{figure}[htbp]
    \centering
    \includegraphics[width=0.6\linewidth,trim=0bp 10bp 0bp 5bp,clip]{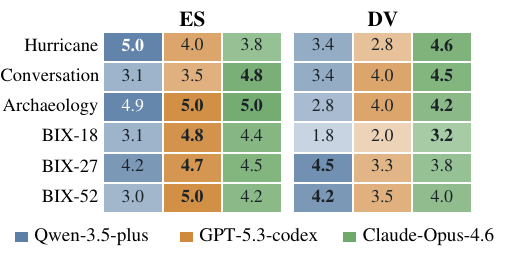}
    \caption{Backbone sensitivity on six shared tasks. Each cell reports the task-level mean ES or DV score for one backbone, using the same judges as in the main evaluation. Rows include two BLADE tasks, one DiscoveryBench task, and three BixBench tasks: BIX-18, BIX-27, and BIX-52. Darker shading indicates higher scores.}
    \label{fig:backbone_results}
\end{figure}

We also test whether the framework depends on a specific backbone by running the same agent design with three models on six shared tasks: two from BLADE, one from DiscoveryBench, and three from BixBench. Figure~\ref{fig:backbone_results} reports this small-scale sensitivity analysis. GPT-5.3-codex and Claude-Opus-4.6 generally improve ES, suggesting stronger evidence alignment. DV is more mixed: Claude-Opus-4.6 achieves the strongest or tied-strongest DV on four tasks, while Qwen3.5-plus remains competitive on two BixBench cases. These results suggest that the framework is not tied to a single backbone: stronger models can improve executable analysis and evidence alignment, but high-value discoveries also require effective exploration choices and hypothesis organization.

\section{Conclusion}
We introduced \textsc{StatefulDiscovery}, an autonomous discovery framework that makes discovery state an explicit control interface for open-ended scientific discovery. The agent maintains observations, hypotheses, evidence, candidate claims, and unresolved frontier directions across discovery rounds, allowing accumulated evidence to shape both what to investigate next and how strongly emerging claims are reported. In open-ended data discovery across 40 real-data tasks, \textsc{StatefulDiscovery} achieves the highest observed overall DV and HQ count/rate, and is preferred in pairwise final claim-set comparisons over raw coding, surprise-guided discovery, evolutionary program search, and objective-evolving optimization baselines. Ablations further show that structured hypotheses and dual-layer architecture contribute to the quality--yield trade-off. These results suggest that explicit discovery state is useful for building scientific agents that produce evidence-calibrated and scientifically valuable claims.

\section*{Acknowledgements}

This work was funded by the New Generation Artificial Intelligence-National Science and Technology Major Project (2026ZD0127802), and the High-level Special Funds (G03050K007) from Southern University of Science and Technology (SUSTech). We thank Jianming Cai for label annotation to validate the LLM-as-a-judge scores. We also thank the reviewers for their comments regarding our work.

\bibliographystyle{plainnat}
\bibliography{references}

\newpage
\appendix
\section{Discovery Trace Example}
\begin{figure*}[h]
    \centering
    \includegraphics[width=\textwidth,trim=7bp 148bp 239bp 0bp,clip]{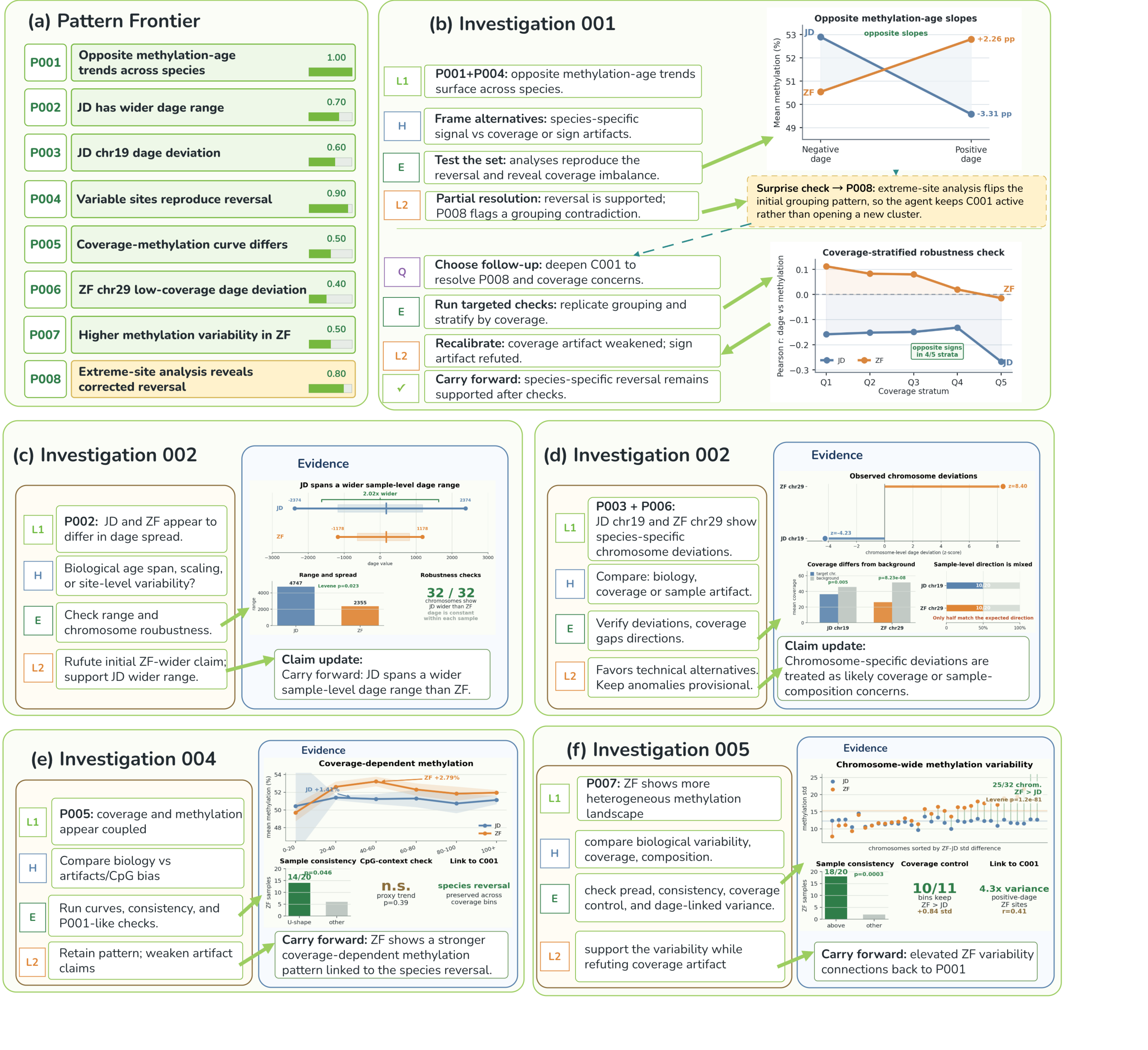}
    \caption{Discovery trace example for the BIX-52 case. The panel shows the prioritized pattern frontier, a detailed L1/L2 trace for Investigation 001, and summary views of subsequent investigations.}
    \label{fig:bix52_complete_case}
\end{figure*}

Figure~\ref{fig:bix52_complete_case} illustrates one complete trace from the BIX-52 task. The example shows how the initial pattern frontier is converted into bounded investigations, how one investigation is decomposed into hypotheses and executable evidence, and how L1/L2 updates determine whether later budget is used to deepen, switch, or retire directions. The figure is intended as a process-level view of the framework.

\label{app:complete_trace_example}
\section{Additional Experimental Details}
\label{app:experimental_details}

\paragraph{Task construction.}
We construct 40 open-ended data-discovery tasks from existing question-answering benchmarks. For each source task, we retain the dataset files and variable descriptions needed to understand and parse the data, including file paths, variable names, column descriptions when available, and data types. We remove the benchmark-specific question, target answer, and concrete analytical query. The resulting task asks an agent to explore the dataset and produce evidence-supported claims without being given a target claim to verify.

\paragraph{Dataset sources.}
The evaluation set draws from three publicly released benchmarks. We do not construct new data or attempt to identify individuals.
\begin{itemize}[leftmargin=*,itemsep=1pt,topsep=1pt,parsep=0.5pt,partopsep=0.5pt]
    \item \textbf{BixBench}~\citep{mitchener2025bixbench} pairs biology questions with associated datasets. We use 21 datasets from BixBench-Verified-50\footnote{\url{https://huggingface.co/datasets/phylobio/BixBench-Verified-50}}, covering genomics, transcriptomics, proteomics, epigenomics, microbiome analysis, and biomedical statistics.
    \item \textbf{BLADE}~\citep{gu2024blade} contains realistic tabular data-analysis tasks. We use 13 datasets spanning social science, education, economics, policy, sports, behavioral data, and synthetic settings.
    \item \textbf{DiscoveryBench}~\citep{majumder2025discoverybench} contains hypothesis-driven discovery tasks across scientific domains. We use 6 datasets covering archaeology, socioeconomic outcomes, education and development indicators, and software-engineering research.
\end{itemize}

\paragraph{Execution environment.}
\textsc{StatefulDiscovery} runs through OpenCode\footnote{\url{https://github.com/opencode-ai/opencode}} in a Docker sandbox. For each task, the sandbox mounts the dataset files and a Markdown instruction file. Appendix~\ref{app:agent_instructions} gives the instruction template used for \textsc{StatefulDiscovery}. All methods are evaluated without web search or external literature retrieval. Agents may use the provided scientific Python environment and install analysis packages when needed. Unless otherwise stated, we use Qwen3.5-plus as the backbone model and set the budget to 40 code executions per task. All executed code, intermediate outputs, structured claims, and final reports are saved for evaluation. 

\paragraph{Artifact licenses and terms.}
We use public datasets, software, and model APIs under their stated licenses or terms. Confirmed public license metadata includes Apache-2.0 for BixBench-Verified-50 and BLADE, MIT for OpenCode, and Apache-2.0 for OpenEvolve. Other third-party artifacts are cited at their source releases or documentation. We do not redistribute third-party datasets, model weights, or baseline code beyond links and reproduction instructions.

\section{Additional Experimental Results}
\subsection{Efficiency and Cost}
\label{app:efficiency_cost}

Table~\ref{tab:efficiency_cost} reports average wall-clock time and API cost per task as practical run-time references. We ran the methods under the same network and API service conditions, so the numbers provide an implementation-level comparison.

\begin{table}[htbp]
\centering
\small
\caption{Average runtime and cost per task. AutoDiscovery uses Qwen3.5-plus as the backbone model. Costs are reported in USD.}
\label{tab:efficiency_cost}
\begin{tabular*}{0.7\linewidth}{@{\extracolsep{\fill}}lcc@{}}
\toprule
\textbf{Method} & \textbf{Cost / task} & \textbf{Time / task} \\
\midrule
AutoDiscovery & \$0.83 & 105 min \\
OpenEvolve & \$0.58 & 57 min \\
SAGA & \$0.59 & 73 min  \\
Raw agent & \$0.22 & 23 min \\
\textsc{StatefulDiscovery} & \$0.55 & 42 min \\
\bottomrule
\end{tabular*}
\end{table}

\subsection{Pairwise Judge Robustness}
\label{app:pairwise_robustness}

As a robustness check, we repeat the pairwise claim-set comparison with GPT-5.5 as an independent judge. Table~\ref{tab:pairwise_judge_robustness} shows the same qualitative direction: \textsc{StatefulDiscovery} wins more often than it loses against every baseline, although the GPT-5.5 judge is more conservative than the Gemini-3.1-pro judge.

\begin{table}[htbp]
\centering
\small
\caption{Pairwise claim-set robustness across two LLM judges. Win/tie/loss counts are from \textsc{StatefulDiscovery}'s perspective over 40 tasks. Gemini-3.1-pro is the primary judge used in the main results. GPT-5.5 is used only for this robustness check.}
\label{tab:pairwise_judge_robustness}
\begin{tabular*}{0.7\linewidth}{@{\extracolsep{\fill}}lcc@{}}
\toprule
Comparison & \makecell{Gemini-3.1-pro\\Win/Tie/Loss} & \makecell{GPT-5.5\\Win/Tie/Loss} \\
\midrule
Ours vs Raw agent & $25/7/8$ & $22/9/9$ \\
Ours vs AutoDiscovery & $32/7/1$ & $30/6/4$ \\
Ours vs OpenEvolve & $40/0/0$ & $37/2/1$ \\
Ours vs SAGA & $31/7/2$ & $27/6/7$ \\
\bottomrule
\end{tabular*}
\end{table}
\subsection{Pairwise Input Length Analysis}
\label{app:pairwise_length}

Prior work has shown that LLM-based preference judges can exhibit verbosity or length bias, often favoring longer outputs~\citep{zheng2023judging,dubois2024length,hu2024explaininglengthbias}. We therefore measure the length of the normalized claim-set inputs used in our pairwise comparisons. The measurement uses the exact formatting function used for pairwise judging: each claim set is represented by claim text plus associated evidence. We report the number of claims per task and the number of formatted characters per task. Table~\ref{tab:pairwise_length_tests} reports paired two-sided Wilcoxon tests against \textsc{StatefulDiscovery}. The descriptive means in Table~\ref{tab:pairwise_length} show the direction of the differences.

\begin{table}[htbp]
\centering
\small
\caption{Length statistics for normalized pairwise-judge inputs over the 40 tasks. Characters are measured after the same formatting and truncation used for pairwise judging.}
\label{tab:pairwise_length}
\begin{tabular*}{0.8\linewidth}{@{\extracolsep{\fill}}lcc@{}}
\toprule
Agent & Claims/task & Chars/task \\
\midrule
\textsc{StatefulDiscovery} & $6.53 \pm 2.71$ & $7365 \pm 3610$ \\
Raw agent & $9.53 \pm 3.07$ & $10480 \pm 3931$ \\
AutoDiscovery & $9.68 \pm 0.76$ & $11488 \pm 996$ \\
OpenEvolve & $8.78 \pm 4.26$ & $10063 \pm 6163$ \\
SAGA & $6.25 \pm 2.13$ & $10756 \pm 4096$ \\
\bottomrule
\end{tabular*}
\end{table}

\begin{table}[htbp]
\centering
\small
\caption{Paired two-sided Wilcoxon tests for normalized pairwise input lengths, comparing each baseline against \textsc{StatefulDiscovery}.}
\label{tab:pairwise_length_tests}
\begin{tabular*}{0.8\linewidth}{@{\extracolsep{\fill}}lcc@{}}
\toprule
Baseline & Claims/task $p$ & Chars/task $p$ \\
\midrule
Raw agent & $9.06{\times}10^{-5}$ & $1.70{\times}10^{-4}$ \\
AutoDiscovery & $1.96{\times}10^{-6}$ & $1.46{\times}10^{-7}$ \\
OpenEvolve & $0.0037$ & $0.0310$ \\
SAGA & $0.624$ & $1.08{\times}10^{-4}$ \\
\bottomrule
\end{tabular*}
\end{table}

\textsc{StatefulDiscovery} uses fewer claims than Raw agent, AutoDiscovery, and OpenEvolve, and a comparable number of claims to SAGA. Its formatted pairwise inputs are significantly shorter than those of all baselines under paired two-sided Wilcoxon tests. This does not rule out all length effects, but it reduces the concern that the pairwise preference results are driven by longer claim-set presentations.

\subsection{Additional Query-Guided Evaluation}
\label{app:query_guided}
To situate the query-free setting, we also evaluate a query-guided variant on two fixed-query benchmarks, BixBench-Verified-50 and HeurekaBench-Lite~\citep{panigrahi2026heurekabench}. We compare four agents in this setting. StatefulDiscovery-QG retains the query decomposition and L2 adjudication skills, while the Raw agent uses the same coding environment and backbone with a direct analysis-and-answer prompt. Crow~\citep{mitchener2025bixbench} is a ReAct-like agent that writes code and observes execution results. Biomni~\citep{huang2025biomni} is a biomedical agent with domain-specific tools and retrieval-augmented know-how. Table~\ref{tab:query_guided_results} shows that StatefulDiscovery-QG performs competitively on both benchmarks, especially on BixBench-Verified-50 tasks that require multi-step tool use and precise numerical answers. The result is consistent with our framing: when a target question is specified, coding agents can often execute and verify the required analysis. The open-ended challenge lies in deciding what scientifically meaningful question to investigate when no target query is given.

\begin{table}[htbp]
    \centering
    \small
    \caption{Query-guided evaluation. BixBench-V50 reports exact/range/LLM-verified answer accuracy on a curated subset of BixBench. HeurekaBench reports G-Eval mean score over open-ended single-cell questions. Biomni excludes one leakage case.}
    \label{tab:query_guided_results}
    \begin{tabular*}{0.7\linewidth}{@{\extracolsep{\fill}}lcc@{}}
        \toprule
        Agent & V50 Acc. $\uparrow$ & Heureka $\uparrow$ \\
        \midrule
        Crow & $70.0\%$ & $3.32$ \\
        Biomni & $60.0\%$ & $\textbf{3.57}$ \\
        Raw agent & $76.0\%$ & $3.36$ \\
        StatefulDiscovery-QG & $\textbf{80.0\%}$ & $3.41$ \\
        \bottomrule
    \end{tabular*}
\end{table}

\section{Baseline Implementation Prompts}
\label{app:baseline_prompts}
Curly-braced fields denote task-specific inputs supplied through the task packet. The same task-specific fields are provided to all coding-agent baselines.

\subsection{Raw Agent Instruction}
The raw agent uses the same dataset files, variable information, coding environment, and code-execution budget as \textsc{StatefulDiscovery}, but receives a direct analysis-and-report instruction without persistent discovery objects, L1 frontier control, or L2 adjudication. The prompt asks the agent to explore the dataset, write executable analyses, test hypotheses with statistical evidence, and produce evidence-linked final claims.

\begin{figure*}[p]
\begin{promptfigurebox}
\footnotesize
\begin{minipage}[t]{0.49\linewidth}
\raggedright
\textbf{Scientific Data Analysis Agent.}
You are a scientific data analysis agent running in an isolated container.
Your goal is to explore a dataset and discover interesting, non-obvious findings.

\vspace{3pt}
\textbf{Task inputs.}
The task packet provides:
\begin{itemize}[leftmargin=1.3em,itemsep=0pt,topsep=1pt,parsep=0pt]
    \item \pvar{dataset\_file\_paths}: paths to dataset files
    \item \pvar{variable information}: column descriptions, variable names, data types, and available dataset context
    \item \pvar{budget.max\_experiments}: maximum number of code executions
\end{itemize}

\vspace{2pt}
\textbf{Task.}
Read the task packet to understand the dataset, then explore the data and discover findings.
\end{minipage}
\hfill
\begin{minipage}[t]{0.47\linewidth}
\raggedright

\textbf{Guidelines.}
\begin{itemize}[leftmargin=1.3em,itemsep=0pt,topsep=1pt,parsep=0pt]
    \item Write and run Python code to analyze the data.
    \item Save each analysis script and its results.
    \item Look for patterns, anomalies, relationships, and unexpected structure.
    \item Test hypotheses with statistical evidence.
    \item The task packet specifies \pvar{budget.max\_experiments}.
    \item You may stop earlier if you judge that the dataset has been sufficiently explored.
    \item You must stop when the code-execution budget is exhausted.
\end{itemize}

\vspace{2pt}
\textbf{Output requirements.}
Write two files before finishing:
\begin{enumerate}[leftmargin=1.5em,itemsep=0pt,topsep=1pt,parsep=0pt]
    \item \texttt{claims.json}: a JSON list of findings, where each claim includes a claim id, description, evidence, experiment references, and confidence score.
    \item \texttt{report.md}: a summary report of the analysis and key findings.
\end{enumerate}
Do not only print results to the console. Final claims must be written to the required output files.
\end{minipage}
\end{promptfigurebox}
\caption{Raw agent instruction template.}
\label{fig:raw_agent_prompt}
\end{figure*}

\begin{figure*}[p]
\begin{promptfigurebox}
\footnotesize
\begin{minipage}[t]{0.49\linewidth}
\raggedright
\textbf{1. Direction proposal.}\\
You are preparing or evolving a discovery program for a data-driven scientific discovery task. Choose one concrete, executable direction for the next experiment.

\textbf{Inputs.} \pvar{dataset\_file\_paths}: paths to dataset files; \pvar{variable information}: column descriptions, variable names, data types, and available dataset context; \pvar{budget.max\_experiments}: maximum number of code executions; program scaffold or current program; optional metrics, improvement areas, and evolution history.\\
\textbf{Return strict JSON.} \texttt{analysis\_type}, \texttt{target\_files}, \texttt{hypothesis}, and \texttt{variables}.\\
\textbf{Rules.} Use only files and variables in the task context; respect file formats; avoid identifier, row-order, sample-number, or coordinate effects unless essential.

\vspace{4pt}
\textbf{2. Direction-conditioned program generation / mutation.}\\
You are improving a discovery program, not writing a general-purpose solver. Modify the program so it executes the selected discovery direction.

\textbf{Inputs.} \pvar{dataset\_file\_paths}: paths to dataset files; \pvar{variable information}: column descriptions, variable names, data types, and available dataset context; \pvar{budget.max\_experiments}: maximum number of code executions; selected direction: \pvar{selected\_direction}; program scaffold or current program; optional metrics and evolution history.\\
\textbf{Task.} Preserve task input loading, the required discovery output schema, and traceability between computed results and emitted claims. Update the direction, analysis, and claim-building logic when applicable.\\
\textbf{Rules.} Use exact data paths; choose parsing from variable information and filename extensions; prefer installed scientific Python libraries; reuse helpers; do not rewrite unrelated scaffolding.\\
\textbf{Return.} SEARCH/REPLACE diffs only.
\end{minipage}
\hfill
\begin{minipage}[t]{0.47\linewidth}
\raggedright

\textbf{3. Scalar fitness judging.}\\
You are a careful judge for data-driven scientific discovery. Evaluate each candidate claim on two axes.

\textbf{Inputs.} \pvar{dataset\_file\_paths}: paths to dataset files; \pvar{variable information}: column descriptions, variable names, data types, and available dataset context; \pvar{budget.max\_experiments}: maximum number of code executions; baseline initial analysis: \pvar{baseline\_initial\_output}; candidate program output: \pvar{candidate\_output}; candidate claims: \pvar{candidate\_claims}.\\
\textbf{Axes.}
\begin{itemize}[leftmargin=1.3em,itemsep=0pt,topsep=1pt,parsep=0pt]
    \item \textbf{alignment}: whether the claim is supported by the candidate program output
    \item \textbf{non-triviality}: whether the claim goes beyond the baseline initial analysis
\end{itemize}
\textbf{Scoring rules.} Score alignment low if the claim overstates, contradicts, or is not supported by the output. Score non-triviality low for descriptive statistics, simple correlations, obvious EDA findings, or findings already revealed by the baseline. Prefer non-obvious relationships, interactions, subgroup effects, or structure.\\
\textbf{Return JSON only.} One record per claim with \texttt{claim\_id}, \texttt{alignment\_bucket}, \texttt{non\_triviality\_bucket}, and short \texttt{reasoning}.
\end{minipage}
\end{promptfigurebox}
\caption{OpenEvolve adaptation prompt snippets.}
\label{fig:openevolve_prompt}
\end{figure*}

\begin{figure*}[t]
\begin{promptfigurebox}
\footnotesize
\textbf{High-level discovery goal.}
Explore the provided dataset and produce novel, evidence-backed scientific findings under the given exploration and experiment budget.

\vspace{3pt}
\textbf{Task inputs.}
\pvar{dataset\_file\_paths}: paths to dataset files; \pvar{variable information}: column descriptions, variable names, data types, and available dataset context; \pvar{budget.max\_experiments}: maximum number of code executions.

\vspace{3pt}
\textbf{Analysis-generation inputs.}
High-level discovery goal; current SAGA objectives; score-selected parent findings from the archive; task inputs; analysis identifier.

\vspace{3pt}
\textbf{Task.}
Return a JSON object containing a brief rationale and one standalone Python script for an executable dataset analysis.

\vspace{3pt}
\textbf{Script requirements.}
Read only from the provided dataset directory, run in the local Python environment, and write a JSON list of findings to the specified output path.
Each finding must contain a claim, evidence, data references, and limitations.
For statistical findings, computed statistics such as effect sizes, test statistics, $p$-values, sample sizes, confidence intervals, and methods should appear in the claim or evidence.

\vspace{3pt}
\textbf{Restrictions.}
Do not use web search, external knowledge bases, external datasets, package installation, or writes to the dataset directory.
Temporary or derived files may only be written to the analysis workspace or temporary directories.

\vspace{3pt}
\textbf{Candidate schema.}
\begin{Verbatim}[breaklines=true,breakanywhere=true,fontsize=\scriptsize]
{
  "claim": string,
  "evidence": string or object,
  "execution_output": object,
  "data_references": {"files": list, "columns": list},
  "limitations": string
}
\end{Verbatim}
Claims are required to be supported by the computed evidence.
\end{promptfigurebox}
\caption{SAGA adaptation prompt template.}
\label{fig:saga_prompt}
\end{figure*}

This baseline is therefore not an intentionally under-specified prompt baseline: it is explicitly instructed to search for non-obvious findings and support them with executable statistical evidence. The difference is that it must manage exploration, evidence status, and claim formation implicitly in the model context rather than through the externalized state used by \textsc{StatefulDiscovery}.

\subsection{OpenEvolve Adaptation Prompts}
OpenEvolve is adapted as an evolutionary program-search baseline. We retain its population-based mutation and scalar selection structure, but condition program mutations on explicit discovery directions so that candidate programs remain aligned with open-ended data discovery. Figure~\ref{fig:openevolve_prompt} shows cleaned prompt snippets for the direction proposal, direction-conditioned mutation, and scalar fitness judging stages.

\subsection{SAGA Adaptation Prompt}
We keep SAGA's planner, scorer-construction, and analyzer prompt structure unchanged, and instantiate its candidate-generation interface for open-ended dataset discovery. Each run provides SAGA with the dataset, exploration budget, current objectives, and high-level goal shown in Figure~\ref{fig:saga_prompt}.

\begin{figure*}[b]
\begin{promptfigurebox}
\footnotesize
\textbf{System.}
You are an expert scientific evaluator. Your task is to assess whether the experimental evidence supports a given claim. Be rigorous: check specific numbers, statistical tests, and reasoning logic.

\vspace{3pt}
\textbf{User inputs.}
\begin{itemize}[leftmargin=1.3em,itemsep=0pt,topsep=1pt,parsep=0pt]
    \item \textbf{Dataset:} \pvar{dataset\_description}
    \item \textbf{Claim:} \pvar{claim\_text}
    \item \textbf{Experimental evidence:} \pvar{code\_output}
\end{itemize}

\vspace{3pt}
\textbf{Question.}
Does the experimental evidence support this claim? Judge the logical soundness of the inference from evidence to conclusion.

\vspace{3pt}
\textbf{Rubric.}
\begin{itemize}[leftmargin=1.3em,itemsep=0pt,topsep=1pt,parsep=0pt]
    \item \textbf{5:} Evidence strongly supports the claim. The inference from data to conclusion is logically sound and statistically backed.
    \item \textbf{4:} Core conclusion is well-supported. Minor gaps do not undermine the main finding.
    \item \textbf{3:} Evidence supports the general direction, but some inferences are only partially justified by the data.
    \item \textbf{2:} Evidence is weak or the claim significantly overstates what the data shows.
    \item \textbf{1:} The claim contradicts the evidence, or there is no evidence provided.
\end{itemize}

\vspace{3pt}
\textbf{Return only JSON:} \texttt{evidential\_support} as an integer from 1 to 5 and \texttt{evidential\_support\_rationale} in 2--3 sentences.
\end{promptfigurebox}
\caption{Evidential-support judging prompt.}
\label{fig:es_prompt}
\end{figure*}

\newpage
\section{Evaluation Prompt Templates}
\label{app:eval_prompts}
We lightly adapt the judge templates used in implementation to match the terminology of this paper. The scoring rubrics are unchanged in substance. The pairwise comparison uses the same position-swapped protocol described in the main text.

\begin{figure*}[p]
\begin{promptfigurebox}
\footnotesize
\textbf{System.}
You are an expert scientific evaluator. Your task is to assess the Discovery Value of the insight established by a candidate scientific claim relative to a provided first-pass analysis.
Assume the claim is factually correct. Do not judge whether the claim is well-supported; evidential validity is evaluated separately.

\vspace{3pt}
\textbf{User inputs.}
\begin{itemize}[leftmargin=1.3em,itemsep=0pt,topsep=1pt,parsep=0pt]
    \item \textbf{Dataset:} \pvar{dataset\_description}
    \item \textbf{First-pass analysis results:} \pvar{baseline\_results}
    \item \textbf{Claim to evaluate:} \pvar{claim\_text}
\end{itemize}

\vspace{3pt}
\textbf{Question.}
Evaluate the Discovery Value of the claim relative to the first-pass results.
Consider novelty relative to the first-pass analysis, explanatory depth, the new understanding established by the claim, and the scope of the insight.

\vspace{3pt}
\textbf{Rubric.}
\begin{itemize}[leftmargin=1.3em,itemsep=0pt,topsep=1pt,parsep=0pt]
    \item \textbf{5:} Very high value. The claim establishes a new organizing interpretation, structural explanation, or data-generating insight that changes how the dataset or phenomenon should be understood overall.
    \item \textbf{4:} High value. The claim establishes an important but more local explanatory insight, such as a confounder, boundary condition, heterogeneity pattern, or organizing relationship within part of the dataset or phenomenon.
    \item \textbf{3:} Moderate value. The claim adds a meaningful refinement beyond first-pass analysis, but mainly concerns a local association, subgroup difference, single analytical comparison, or incremental interpretation.
    \item \textbf{2:} Limited value. The claim is mostly a direct derivative, simple refinement, or obvious implication of the first-pass analysis.
    \item \textbf{1:} Minimal value. The claim is directly readable from the first-pass analysis, restates it, or adds no meaningful new interpretation.
\end{itemize}

\vspace{3pt}
\textbf{Scoring rule.}
Score the insight conveyed by the claim, not the complexity of the method used to obtain it.
Do not reward broad or ambitious wording unless the claim actually provides broad explanatory or structural insight beyond the first-pass analysis.
Treat evidential support, correctness, and reproducibility as separate from this score.

\vspace{3pt}
\textbf{Return only JSON:} \texttt{discovery\_value} as an integer from 1 to 5 and \texttt{discovery\_value\_rationale} in 2--3 sentences.
\end{promptfigurebox}
\caption{Discovery-value judging prompt.}
\label{fig:dv_prompt}
\end{figure*}
\begin{figure*}[t]
\begin{promptfigurebox}
\footnotesize
\textbf{System.}
You are a scientist who has commissioned two independent teams to analyze the same dataset. You are reading their final claim sets and deciding which team produced more valuable scientific insights.

\vspace{3pt}
\textbf{User inputs.}
\begin{itemize}[leftmargin=1.3em,itemsep=0pt,topsep=1pt,parsep=0pt]
    \item \textbf{Dataset:} \pvar{dataset\_description}
    \item \textbf{Team A claim set:} \pvar{system\_a\_claims}
    \item \textbf{Team B claim set:} \pvar{system\_b\_claims}
\end{itemize}

\vspace{3pt}
\textbf{Question.}
As a domain expert, which team's claim set would you find more valuable?

\vspace{3pt}
\textbf{Consider.}
\begin{itemize}[leftmargin=1.3em,itemsep=0pt,topsep=1pt,parsep=0pt]
    \item Does the claim set tell you something you would not have known from basic summary statistics?
    \item Are the conclusions supported by rigorous evidence, or are they speculative?
    \item Does the analysis reveal underlying mechanisms, confounders, or non-obvious relationships?
    \item Are there overreaching or contradictory claims?
\end{itemize}
Do \textbf{not} favor a claim set simply because it has more findings.

\vspace{3pt}
\textbf{Return only JSON:} \texttt{winner} as \texttt{A}, \texttt{B}, or \texttt{tie}, and \texttt{rationale} in 2--3 sentences.
\end{promptfigurebox}
\caption{Pairwise claim-set comparison prompt.}
\label{fig:pairwise_prompt}
\end{figure*}

\clearpage
\section{Human Annotation Instructions}
\label{app:human_annotation}
Annotators were non-author research-group colleagues, with annotation covered by existing research/service responsibilities and compensation arrangements.

\subsection{Discovery Value}
Human annotators used the same Discovery Value (DV) rubric as the automatic judge, rendered in a human-facing annotation form. Figure~\ref{fig:dv_annotation_instruction} reports the instructions given to annotators.

\begin{figure*}[htbp]
\begin{promptfigurebox}
\footnotesize
\begin{minipage}[htbp]{0.48\linewidth}
\raggedright
\textbf{Task.}
Evaluate the Discovery Value (DV) of a candidate scientific claim relative to a provided first-pass analysis.
Assume the claim is factually correct.
Do not judge whether the claim is well-supported, statistically reliable, or reproducible; evidential validity is evaluated separately.

\vspace{3pt}
\textbf{Annotation inputs.}
\begin{itemize}[leftmargin=1.3em,itemsep=1pt,topsep=1pt,parsep=0pt]
    \item Dataset description.
    \item First-pass analysis results.
    \item Claim to evaluate.
\end{itemize}

\vspace{3pt}
\textbf{Question.}
How much new scientific or analytical understanding does the claim establish beyond the first-pass analysis?
Consider novelty relative to the first-pass analysis, explanatory depth, the new understanding established by the claim, and the scope of the insight.

\vspace{3pt}
\textbf{Scoring rule.}
Score the insight conveyed by the claim, not the complexity of the method used to obtain it.
Do not reward broad or ambitious wording unless the claim actually provides broad explanatory or structural insight beyond the first-pass analysis.
Treat evidential support, correctness, and reproducibility as separate from this score.
\end{minipage}
\hfill
\begin{minipage}[htbp]{0.48\linewidth}
\raggedright
\textbf{Rubric.}
\begin{itemize}[leftmargin=1.3em,itemsep=1pt,topsep=1pt,parsep=0pt]
    \item \textbf{5: Very high value.} The claim establishes a new organizing interpretation, structural explanation, or data-generating insight that changes how the dataset or phenomenon should be understood overall.
    \item \textbf{4: High value.} The claim establishes an important but more local explanatory insight, such as a confounder, boundary condition, heterogeneity pattern, or organizing relationship within part of the dataset or phenomenon.
    \item \textbf{3: Moderate value.} The claim adds a meaningful refinement beyond first-pass analysis, but mainly concerns a local association, subgroup difference, single analytical comparison, or incremental interpretation.
    \item \textbf{2: Limited value.} The claim is mostly a direct derivative, simple refinement, or obvious implication of the first-pass analysis.
    \item \textbf{1: Minimal value.} The claim is directly readable from the first-pass analysis, restates it, or adds no meaningful new interpretation.
\end{itemize}

\vspace{3pt}
\textbf{Annotation output.}
Annotators provided an integer \texttt{dv\_score} from 1 to 5.
They could mark \texttt{uncertain} when the claim could not be confidently scored, and could provide an optional one- to two-sentence rationale.
\end{minipage}
\end{promptfigurebox}
\caption{Discovery Value human annotation instructions.}
\label{fig:dv_annotation_instruction}
\end{figure*}

\subsection{Evidential Support}
Human annotators used the same Evidential Support (ES) rubric as the automatic judge, rendered in a human-facing annotation form. Figure~\ref{fig:es_annotation_instruction} reports the instructions given to annotators.

\begin{figure*}[tb]
\begin{promptfigurebox}
\footnotesize
\begin{minipage}[t]{0.48\linewidth}
\raggedright
\textbf{Task.}
Evaluate whether the provided experimental evidence supports a given claim.
Judge only evidence-to-claim alignment.
Do not evaluate whether the claim is novel, important, or high in Discovery Value.

\vspace{3pt}
\textbf{Annotation inputs.}
\begin{itemize}[leftmargin=1.3em,itemsep=1pt,topsep=1pt,parsep=0pt]
    \item Dataset description.
    \item Claim to evaluate.
    \item Experimental evidence, such as code output, statistical results, tables, or summaries.
\end{itemize}

\vspace{3pt}
\textbf{Question.}
Is the given experimental evidence sufficient to support the claim?
Check whether the claim's main direction matches the evidence, whether numbers, variables, groups, and statistical tests are grounded in the evidence, and whether the evidence covers the claim's stated scope.

\vspace{3pt}
\textbf{Scoring rule.}
Do not penalize a claim for low discovery value or limited scientific importance if it is well-supported by the evidence.
Penalize claims that overstate the evidence, turn association into causation, generalize beyond the tested scope, ignore important negative or nonsignificant results, or report numbers, variables, or directions that conflict with the evidence.
\end{minipage}
\hfill
\begin{minipage}[t]{0.48\linewidth}
\raggedright
\textbf{Rubric.}
\begin{itemize}[leftmargin=1.3em,itemsep=1pt,topsep=1pt,parsep=0pt]
    \item \textbf{5: Strong support.} The experimental evidence directly and sufficiently supports the claim. Key numbers, statistical tests, group comparisons, and the reasoning chain match the claim, and the claim does not noticeably exceed the evidence.
    \item \textbf{4: Core conclusion supported.} The claim's main conclusion is supported, but there are minor gaps or slightly strong wording that do not undermine the main finding.
    \item \textbf{3: Partial support.} The evidence supports the general direction, but the claim includes some extension, broader scope, or added interpretation that is only partially justified.
    \item \textbf{2: Weak support or clear overinterpretation.} The evidence contains some related signal, but it is insufficient for the main strength, scope, or causal/mechanistic wording of the claim.
    \item \textbf{1: Unsupported or contradicted.} The claim's core content is not tested, lacks direct evidence, reports inconsistent numbers or variables, or contradicts the provided evidence.
\end{itemize}

\vspace{3pt}
\textbf{Annotation output.}
Annotators provided an integer \texttt{es\_score} from 1 to 5.
They could mark \texttt{uncertain} when the evidence was difficult to interpret, and could provide an optional one- to two-sentence rationale.
\end{minipage}
\end{promptfigurebox}
\caption{Evidential Support human annotation instructions.}
\label{fig:es_annotation_instruction}
\end{figure*}

\FloatBarrier
\section{Control Skill Interfaces and Schemas}
\label{app:skill_all}

\subsection{Skill Interfaces}
\label{app:skill_interfaces}

Table~\ref{tab:scaffold_skills} maps loop locations in Figure~\ref{fig:method_overview} to the concrete skills the agent can invoke. Artifact-writing skills maintain persistent objects; frontier-control and adjudication skills support L1 and L2; and decomposition and experiment-running skills support Steps 1--3. Beyond framework-level skills, the experiment runner can invoke specialized domain recipes, such as biomedical or phylogenetic analyses, when a query requires professional methods.

\begin{table}[H]
    \centering
    \footnotesize
    \setlength{\tabcolsep}{3pt}
    \renewcommand{\arraystretch}{1.16}
    \caption{Skills may be invoked at each loop location.}
    \label{tab:scaffold_skills}
    \begin{tabular}{>{\raggedright\arraybackslash}p{0.25\linewidth}>{\raggedright\arraybackslash}p{0.66\linewidth}}
        \toprule
        \textbf{Loop location} & \textbf{Concrete skills} \\
        \midrule
        Initialization &
        \skill{write-patterns}: record candidate patterns and priorities.\newline
        \skill{write-state}: seed the initial frontier. \\
        L1 frontier control &
        \skill{exploration-strategist}: choose create, attach, deepen, switch, retire, or stop.\newline
        \skill{write-investigation}, \skill{write-state}: apply the selected action. \\
        Steps 1--2 &
        \skill{investigation-decomposition}: maintain the hypothesis set and query bundle.\newline
        \skill{write-investigation}: record active requirements. \\
        Step 3 &
        \skill{experiment-runner}: execute analyses.\newline
        \skill{write-experiment}: record outputs as evidence.\\
        Step 4 / L2 adjudication &
        \skill{evidence-strength-judge}: assess support, confidence, red flags, and next-step signals.\newline
        \skill{write-investigation}, \skill{write-claims}, \skill{repro-gate-check}, \skill{write-state}: update hypothesis status, finalize claims, check reproducibility, and update state. \\
        \bottomrule
    \end{tabular}
\end{table}

\subsection{Write Interfaces and State Fields}
\begin{figure*}[htbp]
\begin{promptfigurebox}
\scriptsize
\begin{minipage}[htbp]{0.48\linewidth}
\raggedright
\textbf{name:} \skill{write-patterns}\\
\textbf{description:} Defines \texttt{patterns.json} for tracking unexplained data patterns.

\vspace{2pt}
\textbf{\# Format.}\\
Write to \texttt{patterns.json}; updated throughout exploration.

\vspace{2pt}
\textbf{Schema.}
\begin{Verbatim}[fontsize=\tiny,breaklines=true,breakanywhere=true]
{
  "patterns": [
    {
      "id": "P1",
      "observation": "specific pattern with data values or statistics",
      "source": "eda | base002 | exp003",
      "status": "unexplained | partially_explained | explained",
      "cluster_id": "C1 or null",
      "related_variables": ["column_name"],
      "priority": 0.0
    }
  ]
}
\end{Verbatim}

\vspace{3pt}
\textbf{name:} \skill{write-investigation}\\
\textbf{description:} Defines investigation records, hypothesis sets, and query bundles.

\vspace{2pt}
\textbf{\# Format.}\\
Write to \texttt{investigations.json} and \texttt{current\_investigation.json}; stores all investigations and the active one.

\vspace{2pt}
\textbf{Schema.}
\begin{Verbatim}[fontsize=\tiny,breaklines=true,breakanywhere=true]
{
  "active_investigation_id": "I1",
  "investigations": [
    {
      "id": "I1",
      "phenomenon": "observed phenomenon",
      "pattern_ids": ["P1"],
      "status": "active | paused | resolved | retired | saturated",
      "hypotheses": [
        {
          "id": "I1-h1",
          "role": "main | alternative_explanation | artifact_check | robustness",
          "status": "pending | supported | weakened | refuted | inconclusive",
          "confidence": null,
          "evidence_refs": ["exp003"],
          "remaining_concerns": []
        }
      ],
      "query_bundle": [
        {
          "id": "I1-q1",
          "hypothesis_id": "I1-h1",
          "result_ref": null,
          "status": "pending | done | failed"
        }
      ]
    }
  ]
}
\end{Verbatim}
\end{minipage}
\hfill
\begin{minipage}[htbp]{0.48\linewidth}
\raggedright
\textbf{name:} \skill{write-experiment}\\
\textbf{description:} Defines executable experiment records in \texttt{experiments.jsonl}.

\vspace{2pt}
\textbf{\# Format.}\\
Append one JSON object per query; each record is one budget unit written by the forked \skill{experiment-runner}.

\vspace{2pt}
\textbf{Schema.}
\begin{Verbatim}[fontsize=\tiny,breaklines=true,breakanywhere=true]
{
  "experiment_id": "base001 | exp001",
  "phase": "base | exploration",
  "investigation_id": "I1 or null",
  "query_id": "I1-q1 or null",
  "question": "research question tested",
  "status": "success | failed | inconclusive",
  "summary": "result with key numbers",
  "code_path": "code/exp001.py",
  "artifacts": ["code/exp001_results.json"],
  "hypothesis_refs": ["I1-h1"],
  "limitations": []
}
\end{Verbatim}

\vspace{3pt}
\textbf{name:} \skill{write-state}\\
\textbf{description:} Defines \texttt{epistemic\_state.json} with L1/L2 assessment.

\vspace{2pt}
\textbf{\# Format.}\\
Write to \texttt{epistemic\_state.json}; records frontier and resolution state.

\vspace{2pt}
\textbf{Schema.}
\begin{Verbatim}[fontsize=\tiny,breaklines=true,breakanywhere=true]
{
  "exploration_progress": {
    "cycle_id": 3,
    "budget_remaining": 22
  },
  "level1": {
    "frontier_status": "open | narrowing | exhausted",
    "saturation_assessment": "low | medium | high",
    "recommended_mode": "branch | deepen | switch | stop"
  },
  "level2": {
    "active_investigation_id": "I1",
    "resolution_status": "open | partially_resolved | resolved | inconclusive",
    "resolution_confidence": 0.0,
    "evidence_strength": "weak | moderate | strong",
    "red_flag_pressure": "low | medium | high",
    "next_best_resolution_step": "single uncertainty-reducing action"
  },
  "decision_log": [{"cycle_id": 3, "action": "deepen"}]
}
\end{Verbatim}

\vspace{3pt}
\textbf{name:} \skill{write-claims}\\
\textbf{description:} Defines \texttt{claims.json} for aggregated final claims.

\vspace{2pt}
\textbf{\# Format.}\\
Write to \texttt{claims.json}; updated after L2 adjudication and finalization.

\vspace{2pt}
\textbf{Schema.}
\begin{Verbatim}[fontsize=\tiny,breaklines=true,breakanywhere=true]
{
  "claims": [
    {
      "claim_id": "I1-h1",
      "investigation_id": "I1",
      "role": "main | alternative_explanation | artifact_check | robustness",
      "text": "claim statement with key numbers",
      "status": "supported | weakened | refuted | inconclusive",
      "confidence": 0.0,
      "evidence_refs": ["exp003"],
      "supporting_stats": {}
    }
  ]
}
\end{Verbatim}

\vspace{3pt}
\textbf{Control records.}
\texttt{local\_adjudication} records hypothesis evaluations, confidence, red flags, and next step.
\texttt{strategy\_recommendation} records selected action, target, rationale, status updates, and budget assessment.
\end{minipage}
\end{promptfigurebox}
\caption{Write interfaces and state-field schema used by the framework skills.}
\label{fig:write_interfaces_prompt}
\end{figure*}

\begin{figure*}[t]
\begin{promptfigurebox}
\footnotesize
\textbf{name:} \skill{evidence-strength-judge}\\
\textbf{description:} Invoke after \skill{experiment-runner} completes an investigation query bundle. Evaluates hypotheses, produces a local adjudication, and writes L2 state updates.

\vspace{2pt}
\textbf{\# Epistemic Evidence Assessor.}\\
\textbf{Inputs.}
Read the active investigation, its hypothesis set and query bundle, the referenced experiment records, and the executed code/results needed to verify the evidence.

\vspace{3pt}
\textbf{Checks.}
\begin{enumerate}[leftmargin=1.5em,itemsep=0pt,topsep=1pt,parsep=0pt]
    \item \textbf{Evidence sanity:} verify sample sizes, filters, key statistics, and reported numbers; flag mismatches between code outputs and evidence records.
    \item \textbf{Method fit:} check that code implements the query requirements; flag unauthorized filtering, transformations, or inconsistent preparation.
    \item \textbf{Cross-hypothesis adjudication:} compare the main hypothesis against alternatives; use artifact checks and robustness checks to weaken unsupported explanations.
    \item \textbf{Confidence calibration:} assign status and confidence for each hypothesis; cap confidence when red flags are present.
\end{enumerate}

\vspace{3pt}
\textbf{Decision rules.}
Mark a hypothesis as supported when evidence matches the hypothesis, sanity and method checks pass, and alternatives or artifact explanations are weaker.
Mark it as weakened when evidence is partial, sensitive, or better explained by an alternative or artifact.
Mark it as refuted when evidence contradicts the hypothesis or a competing explanation is clearly better supported.
Mark it as inconclusive when evidence is insufficient, flawed, or ambiguous.
Any red flag caps confidence at $0.6$; unresolved useful gaps trigger deepening; methodological flaws trigger self-correction.

\vspace{3pt}
\textbf{Confidence rubric.}
$0.9+$ means strong evidence with no red flags and robust checks; $0.7$--$0.9$ means reasonable evidence with minor concerns; $0.5$--$0.7$ means uncertain evidence, red flags, or sensitivity to choices; below $0.5$ means weak evidence.

\vspace{3pt}
\textbf{Writes.}
Write a local adjudication control record, hypothesis status and concerns via \skill{write-investigation}, the L2 update to \texttt{epistemic\_state} via \skill{write-state}, and finalizable supported claims via \skill{write-claims}.
\end{promptfigurebox}
\caption{L2 evidence-adjudication skill prompt.}
\label{fig:l2_skill_prompt}
\end{figure*}

\begin{figure*}[t]
\begin{promptfigurebox}
\footnotesize
\textbf{name:} \skill{exploration-strategist}\\
\textbf{description:} Invoke after L2 adjudication or initialization. Recommends the next frontier action by combining pattern priority, investigation status, L2 adjudication signals, and remaining budget.

\vspace{2pt}
\textbf{\# Exploration Strategist.}\\
\textbf{Inputs.}
Read task budget, L1/L2 fields in \texttt{epistemic\_state}, pattern priorities and statuses, investigation statuses, and the latest local adjudication.

\vspace{3pt}
\textbf{Decision policy.}
\begin{enumerate}[leftmargin=1.5em,itemsep=0pt,topsep=1pt,parsep=0pt]
    \item Treat L2 as the freshest signal for the active investigation: resolved high-confidence investigations can be left behind, unresolved useful gaps should be deepened, and high red-flag pressure triggers self-correction, deepening, or retirement.
    \item Assess the frontier: high-priority unexplained patterns create or attach investigations, related patterns are attached to existing investigations, and saturated or low-value residue triggers switching, retirement, or stopping.
    \item Apply budget awareness: low remaining budget favors finishing active investigations, while exhausted budget requires stopping.
\end{enumerate}

\vspace{3pt}
\textbf{Action matrix.}
For an open frontier and resolved investigation, create a new investigation or attach a pattern.
For an open frontier and unresolved investigation, deepen the investigation or create a new one when priorities and budget favor it.
For a narrowing frontier, create a new investigation only if high-priority patterns remain; otherwise deepen unresolved work or stop.
For an exhausted frontier, stop when resolved, or deepen only if unresolved work and budget remain.

\vspace{3pt}
\textbf{Actions.}
The strategy may select \texttt{create\_investigation}, \texttt{attach\_pattern}, \texttt{deepen\_investigation}, \texttt{switch\_investigation}, \texttt{retire\_investigation}, or \texttt{stop}.

\vspace{3pt}
\textbf{Writes.}
Write a \texttt{strategy\_recommendation} control record, pattern status and priority updates via \skill{write-patterns}, investigation action/status updates via \skill{write-investigation}, and the L1 update to \texttt{epistemic\_state} via \skill{write-state}.
\end{promptfigurebox}
\caption{L1 frontier-control skill prompt.}
\label{fig:l1_skill_prompt}
\end{figure*}

\begin{figure*}[t]
    \centering
    \includegraphics[width=0.98\textwidth]{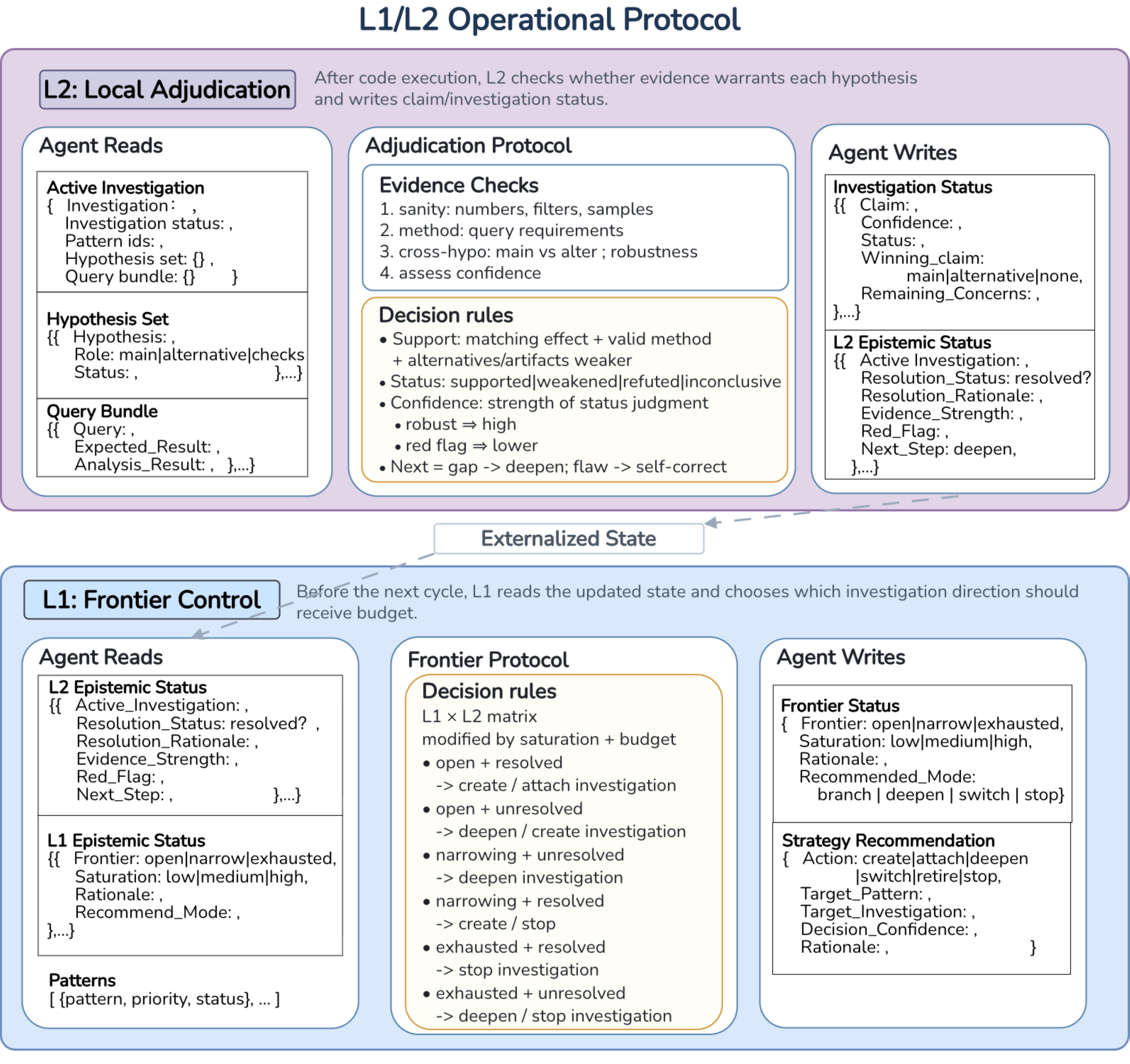}
    \caption{Expanded view of the L1/L2 operational protocol. The upper panel details how L2 reads the active investigation, checks executable evidence, and writes investigation status and L2 epistemic status. The lower panel details how L1 reads the updated state, combines resolution and frontier signals, and writes the next frontier recommendation.}
    \label{fig:l1_l2_schema}
\end{figure*}

Figure~\ref{fig:l1_l2_schema} expands the operational view in the L1 (\skill{exploration-strategist}) and L2\\(\skill{evidence-strength-judge}) skills by showing the state artifacts read and written by each level. We then record the implementation-facing fields used by the corresponding write and decision skills. Figures~\ref{fig:l2_skill_prompt} and~\ref{fig:l1_skill_prompt} provide the corresponding skill prompts.

\section{Agent Instruction Example}
\label{app:agent_instructions}
The task instructions are implemented as a Markdown agent-instruction file mounted in the sandbox. The released repository contains the exact file used in our experiments. Figures~\ref{fig:agent_instruction_init}--\ref{fig:agent_instruction_loop} provide a cleaned example in the same prompt-template style, preserving the execution flow while omitting path-level and provider-specific details.

\begin{figure*}[!t]
\begin{promptfigurebox}
\footnotesize
\textbf{\textsc{StatefulDiscovery} agent instructions: role, inputs, and initialization.}

\vspace{3pt}
You are an autonomous scientific discovery agent running in an isolated sandbox.
Given task instructions, dataset files, and a finite experiment budget, explore the dataset and produce executable-evidence-supported scientific findings.

\vspace{3pt}
\textbf{Task inputs.}
The task packet provides:
\begin{itemize}[leftmargin=1.3em,itemsep=0pt,topsep=1pt,parsep=0pt]
    \item \pvar{dataset\_file\_paths}: paths to dataset files
    \item \pvar{variable information}: column descriptions, variable names, data types, and available dataset context
    \item \pvar{budget.max\_experiments}: maximum number of code executions
\end{itemize}

\vspace{3pt}
\textbf{Persistent artifacts.}
Maintain persistent artifacts for candidate patterns and priorities, investigations and the active investigation, query requirements and evidence records, L2 adjudication records, the externalized epistemic state, L1 strategy recommendations, final claims, and the final report.

\vspace{3pt}
\textbf{Phase 0: Initialization.}
\begin{enumerate}[leftmargin=1.5em,itemsep=0pt,topsep=1pt,parsep=0pt]
    \item Read the task instructions and dataset files.
    \item Run exploratory analyses and cheap base experiments.
    \item Extract data-anchored patterns; each pattern must reference evidence from base analyses and receive an agent-assigned relative priority score in $[0,1]$ among the current patterns.
    \item Record patterns with \skill{write-patterns}.
    \item Initialize the externalized state with \skill{write-state}: frontier status open, saturation low, and no active investigation.
    \item Invoke \skill{exploration-strategist} to select the first investigation direction.
\end{enumerate}
\end{promptfigurebox}
\caption{StatefulDiscovery agent instruction template: role, inputs, and initialization.}
\label{fig:agent_instruction_init}
\end{figure*}

\begin{figure*}[!t]
\begin{promptfigurebox}
\footnotesize
\textbf{StatefulDiscovery agent instructions: exploration loop.}

\vspace{3pt}
Each cycle follows the same rhythm.

\vspace{3pt}
\textbf{1. Read the current strategy recommendation.}
Apply any pattern or investigation-status updates, then execute one frontier action: create investigation, attach pattern to an existing investigation, deepen investigation, switch investigation, retire investigation, or stop.

\vspace{3pt}
\textbf{2. Maintain the active investigation.}
Invoke \skill{investigation-decomposition} to maintain the structured hypothesis set, main hypothesis, alternative hypotheses, artifact checks, robustness checks, and executable query bundle.
Persist the active investigation and query bundle with \skill{write-investigation}.

\vspace{3pt}
\textbf{3. Execute analyses.}
Invoke the forked \skill{experiment-runner} to execute the query bundle.
Record code outputs and evidence records with \skill{write-experiment}.

\vspace{3pt}
\textbf{4. Apply L2 local adjudication.}
Invoke \skill{evidence-strength-judge} to evaluate the active investigation.
For each hypothesis, assign one status: supported, weakened, refuted, or inconclusive.
Update the active investigation with hypothesis status, confidence scores, evidence links and counter-evidence, red flags, and remaining concerns.
Update the externalized state with resolution status, resolution confidence, evidence strength, red-flag pressure, and the next best resolution step.
If sanity or method checks fail, revise the active investigation before advancing the frontier.
Merge finalizable supported claims with \skill{write-claims}.

\vspace{3pt}
\textbf{5. Perform surprise check.}
Compare each query's expected result with its observed result.
If the investigation is resolved and the result is surprising, add a new pattern with \skill{write-patterns} for future frontier control.
If the investigation remains unresolved and the result is surprising, strengthen the case for self-correction or further testing.

\vspace{3pt}
\textbf{6. Apply L1 frontier control.}
Update pattern statuses with \skill{write-patterns}, then invoke \skill{exploration-strategist}.
Choose the next action using the L2 resolution signal, pattern priorities, frontier saturation, red-flag pressure, and remaining budget.
Apply the selected frontier action with \skill{write-investigation} when it creates, attaches, deepens, switches, or retires an investigation.
Update the externalized state with frontier status, saturation, recommended mode, budget progress, and the decision log.
Repeat until the strategy recommendation is stop.

\vspace{3pt}
\textbf{Finalization.}
When the strategy recommendation is stop, aggregate supported claims with \skill{write-claims}, write the final report organized by investigation, and apply \skill{repro-gate-check} to ensure claims link to executable evidence.

\vspace{3pt}
\textbf{Hard constraints.}
\begin{itemize}[leftmargin=1.3em,itemsep=0pt,topsep=1pt,parsep=0pt]
    \item Do not use external search or external data.
    \item Respect \pvar{budget.max\_experiments}; one executed query counts as one experiment.
    \item Every final claim must link to executable evidence.
\end{itemize}

\end{promptfigurebox}
\caption{StatefulDiscovery agent instruction template: the exploration loop.}
\label{fig:agent_instruction_loop}
\end{figure*}

\newpage
\section{Case Study: Surprise and Explicit State}
\label{app:hurricane_case}
To understand the difference behind the aggregated results, we examine the hurricane dataset in BLADE, where the relationship between hurricane-name femininity and deaths can be explained in several ways. Table~\ref{tab:hurricane_case_study} compares AutoDiscovery and StatefulDiscovery on this task.

\begin{table*}[htbp]
    \centering
    \small
    \caption{Case study on the hurricane dataset. For AutoDiscovery, numbers report the means of sampled agreement distributions before and after evidence, together with the resulting surprise/reward $S$; ``down'' indicates that evidence lowers agreement with the original hypothesis.}
    \label{tab:hurricane_case_study}
    \begin{tabular}{p{0.14\linewidth}p{0.40\linewidth}p{0.40\linewidth}}
        \toprule
        Aspect & AutoDiscovery & StatefulDiscovery \\
        \midrule
        Representative outputs
        &
        \begin{minipage}[t]{\linewidth}
        \begin{enumerate}[leftmargin=1.2em,itemsep=0.25em,topsep=0.15em,parsep=0pt]
            \item U-shaped relation between name femininity and property damage ($0.73 \rightarrow 0.26$; down; $S=1.14$).
            \item Low pressure and high femininity jointly increase deaths ($0.27 \rightarrow 0.58$; supported; $S=0.50$).
            \item Quadratic relation between name femininity and deaths, with extreme names deadlier ($0.73 \rightarrow 0.42$; down; $S=0.50$).
            \item Minimum pressure better predicts property damage than wind speed, with category moderation ($0.73 \rightarrow 0.42$; down; $S=0.50$).
        \end{enumerate}
        \end{minipage}
        &
        \begin{minipage}[t]{\linewidth}
        \begin{enumerate}[leftmargin=1.2em,itemsep=0.25em,topsep=0.15em,parsep=0pt]
            \item The gender--death pattern is driven by random clustering of extreme events with female names, not a causal mechanism.
            \item The apparent pattern is a time-period artifact: the 1960s used only female names and included several catastrophic storms.
            \item Binary name gender reflects the historical naming convention; the femininity score is largely redundant with gender.
            \item Category--death associations are confounded by wind speed and inflated by a few high-death outliers.
        \end{enumerate}
        \end{minipage}
        \\
        Organization
        & Separate hypothesis nodes are scored by surprise; a sharp agreement shift can rank a rejected local hypothesis highly.
        & The apparent association is decomposed into linked investigations: gender--death causality vs. outlier/time-period alternatives, masfem naming-convention redundancy, and category/wind/outlier confounding. \\
        \bottomrule
    \end{tabular}
\end{table*}

The two agents discover different kinds of findings. AutoDiscovery ranks hypotheses by surprise, and in most cases the sampled agreement with the original hypothesis decreases after evidence is shown. StatefulDiscovery instead keeps the name--death association as a connected set of investigations, linking gender effects, naming conventions, storm severity, and outlier concentration. This pattern reflects how AutoDiscovery's surprise reward can be high when evidence substantially changes the agreement distribution for a hypothesis, even if the update mainly weakens that hypothesis. Such negative evidence is useful, but it does not by itself determine how related hypotheses should be retained, weakened, or connected. In this sense, Bayesian surprise in AutoDiscovery measures a model-belief update, not necessarily a scientific surprise that changes how the phenomenon is understood.

\end{document}